\PassOptionsToPackage{table}{xcolor}
\documentclass{article}
\usepackage{iclr2027_conference,times}
\usepackage[utf8]{inputenc}
\usepackage[T1]{fontenc}
\usepackage{amsmath,amssymb,booktabs,graphicx,microtype}
\usepackage{url,hyperref}
\usepackage{xcolor}
\usepackage{array,tabularx}
\usepackage[font=small,labelfont=bf,skip=5pt]{caption}
\usepackage{enumitem}
\usepackage{wrapfig}
\usepackage{calc}
\usepackage[breakable]{tcolorbox}
\newtcolorbox{promptbox}[2][]{breakable,colback=black!3,colframe=black!75,
colbacktitle=black!75,coltitle=white,fonttitle=\bfseries,title={#2},
boxrule=0.7pt,arc=2mm,left=9pt,right=9pt,top=8pt,bottom=8pt,
before skip=10pt,after skip=10pt,fontupper=\small,#1}
\providecommand{\promptrole}[1]{\par\smallskip\noindent\textbf{#1}\par\nobreak\smallskip}
\providecommand{\promptvar}[1]{\textit{[#1]}}
\providecommand{\promptnote}[1]{\par\smallskip\noindent\textit{#1}\par\nobreak\smallskip}
\providecommand{\promptsep}{\par\smallskip\noindent\rule{\linewidth}{0.3pt}\par}

\hypersetup{colorlinks=true,linkcolor=blue!45!black,citecolor=blue!45!black,urlcolor=blue!45!black}

\newcommand{\appref}[1]{\hyperref[#1]{Appendix~\ref*{#1}}}
\newcommand{\hl}[1]{\textbf{\textcolor{blue!45!black}{#1}}}

\newcommand{\Abase}{A_{\mathrm{base}}}
\newcommand{\Ako}{A_{\mathrm{KO}}}
\newcommand{\dko}{\Delta_{\mathrm{KO}}}

\iclrfinalcopy  

\title{What Keeps Vision-Language Models\\Looking at the Image?}

\author{
Hiroto Osaka$^{1}$,\,
Shohei Taniguchi$^{1}$,\,
Gouki Minegishi$^{1}$,\\
\,\,\textbf{Kai Yamashita}$^{1}$,\,
\textbf{Masahiro Suzuki}$^{1}$,\,
\textbf{Yutaka Matsuo}$^{1}$
\\[0.25em]
$^{1}$The University of Tokyo
\\[0.25em]
\,\,\texttt{hiroto.osaka@weblab.t.u-tokyo.ac.jp}
}

\begin{document}
\maketitle
\fancyhead[L]{Preprint}

\begin{abstract}
When do vision-language models need direct access to the image while generating an answer?
We study image dependence during answer generation by examining how the visual information needed for the current question becomes available in context.
We intervene on direct image access while retaining previously computed states. 
Across real-image and synthetic tasks, we show that, depending on the generation process, direct access can continue to support accuracy after question processing. 
Supplying the required attributes as text in the context weakens this dependence.
On synthetic tasks, we also examine how dependence changes as the model itself states the required attributes.
Before attribute expression, severing access reduces accuracy, and replacing image-side states shifts answers toward the counterfactual content. 
After sufficient expression, both interventions have smaller effects. 
Even with an identical generated prefix, dependence differs according to whether image access was available during question processing. 
Thus, both the visible text and the preceding image access matter.
Several of these patterns hold across model families, including Qwen2.5-VL-32B and InternVL3-14B. 
These findings offer a view of image dependence in terms of the information needed for the current question and the history of image access, beyond generation position alone. 
This perspective provides a basis for deciding when to reduce visual access during an answer and which visual information to retain for subsequent questions.
\end{abstract}

\section{Introduction}
\label{sec:intro}

% P1 -- question and why it matters (FB07: concrete follow-up example first)
Vision-language models (VLMs) answer questions about images, either directly or through intermediate reasoning.
Chain-of-thought (CoT) methods generate intermediate reasoning before the final answer \citep{wei2022chain, zhang2024multimodal}.
Consider a follow-up question. A model correctly answers ``lower left'' when asked which quadrant contains the only yellow star; when asked next how many green objects that quadrant contains, it needs colors that the first answer never stated.
A correct earlier answer therefore need not contain what the next question needs.
It remains unclear whether the model must read the image again or can rely on what it has already processed and written.
The question bears on when image access can be reduced: efficiency methods reduce visual computation by selecting fewer visual tokens for input to the language model \citep{yang2025visionzip}, pruning visual tokens after an early layer \citep{chen2024fastv}, freezing visual-token updates in selected layers \citep{yuan2025shortv}, or compressing and evicting visual key--value (KV) cache entries \citep{tu2025vlcache, wan2024lookm, li2025madakv}. 
These approaches motivate the complementary question of how sensitivity to continued direct image access changes as answer generation proceeds.
We ask when direct image access is needed during answer generation, focusing on how the visual information required by the current question becomes available in context.
We show that sensitivity to blocking this access varies with the generation process, the attributes available in text, and the preceding image access.

% P2 -- closest prior work and what is open (FB08, FB12)
Prior work has traced how visual information reaches the answer: it flows through direct attention to image tokens and through text-mediated pathways via question tokens \citep{kaduri2025whats, zhang2025crossmodal, yin2025liftingveil, wu2026howvision}, and under some interventions the text-mediated pathway can serve as a fallback when direct image access is disrupted \citep{salazar2026pathways}.
Across dialogue turns, assistant-text states substitute for visual states for facts that were verbalized, less so for facts that were not \citep{chen2026seen}.
During long reasoning, attention to the image fades and visual contributions concentrate early \citep{sun2025tvc, chu2025lookagain, sammani2026scaling}, and re-completing a reasoning chain truncated midway without the image costs little accuracy \citep{sun2025tvc}.
Whether attention weights alone can establish what a model uses has long been debated \citep{jain2019attention, wiegreffe2019attention}, and a fall in attention to the image does not imply a fall in the accuracy cost of severing that attention.
We therefore measure intervention sensitivity directly.
These findings leave open how dependence on direct image access changes as information needed for the current question becomes available during a response.
We address this with three complementary comparisons: question-access contrasts at a fixed visible prefix, knockout aligned to each response's attribute-expression stage, and a directional KV swap before and after that stage (\autoref{sec:discussion}).

\begin{figure}[t]
\centering
\includegraphics[width=0.85\linewidth]{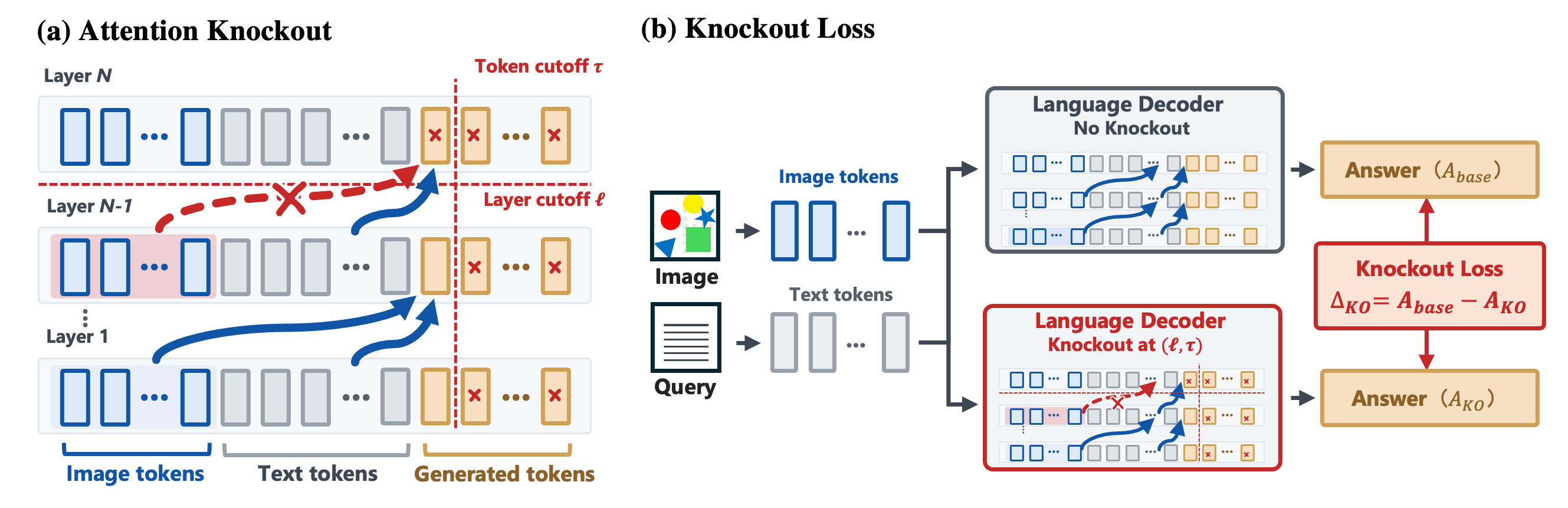}
\caption{\textbf{Attention-knockout intervention and measurement of knockout loss.} \textbf{(a)} Answer-side knockout masks attention to image-token keys when either the layer or the token condition is met ($k\ge\ell$ or $t\ge\tau$). Cached keys and values and untargeted attention edges are retained, while the remaining attention weights are renormalized. Question-side knockout, not shown, targets the current question rows instead. \textbf{(b)} Knockout loss $\dko=\Abase-\Ako$ is the accuracy difference between a matched baseline and the knockout condition, with question-time access held fixed. Layers are numbered from 1 in the schematic and indexed from 0 in the formal definition.}
\label{fig:method}
\end{figure}

\begin{figure}[t]
\centering
\includegraphics[width=0.85\linewidth]{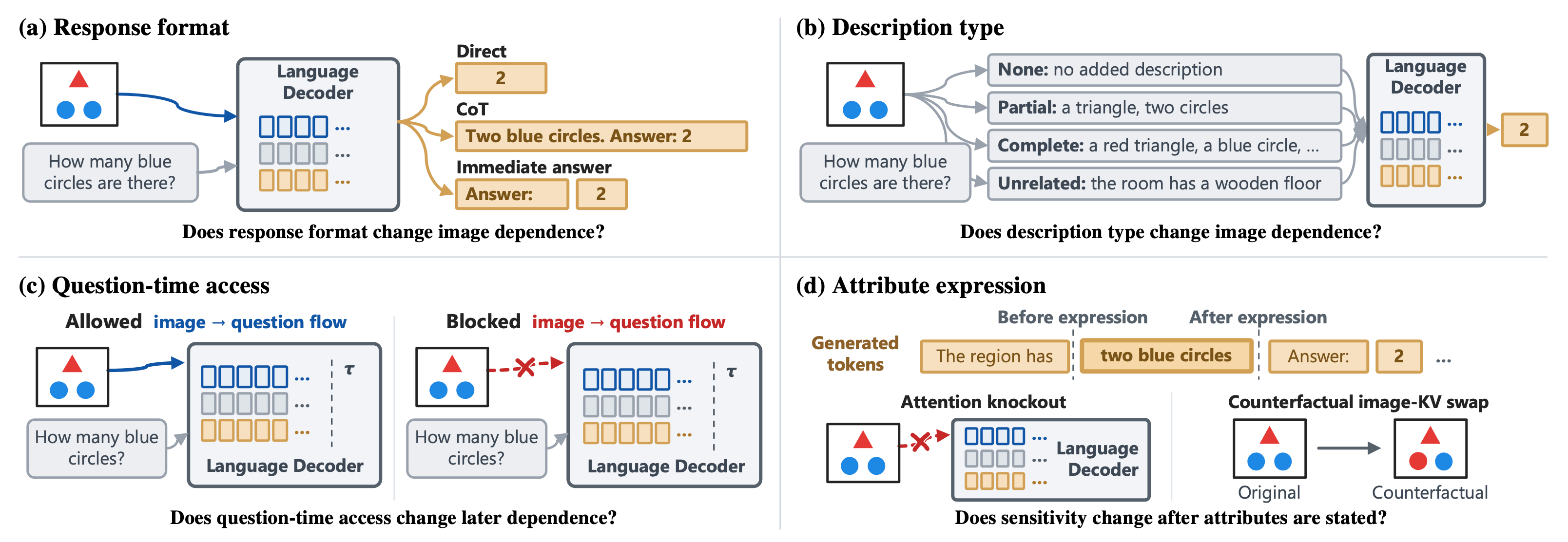}
\caption{\textbf{Experimental design of the four image-dependence comparisons.} \textbf{(a) Response format:} Direct, CoT, and an immediate-answer control that retains the CoT instruction. \textbf{(b) Description type:} whether supplied text contains the attributes needed for the answer. \textbf{(c) Question-time access:} whether the current question can attend to image tokens. \textbf{(d) Attribute expression:} knockout and image-KV swap before and after sufficient attribute expression in the generated response.}
\label{fig:conditions}
\end{figure}

% P3 -- how we test it (FB09, FB10)
We measure dependence with attention knockout \citep{geva2023dissecting, kaduri2025whats}.
As \autoref{fig:method} illustrates, attention from chosen positions to image tokens is severed from a chosen layer and generation step onward, while previously computed states and the remaining attention connections are retained.
The knockout loss, the accuracy drop against a baseline that differs only in the knockout, quantifies this sensitivity.
Within each comparison, we hold the image set and scoring fixed and vary an input condition, access condition, or intervention stage.
Synthetic counting tasks let us control which attributes are supplied in text and when image access is available.
On real images, GQA tests the description and question-time access comparisons, and TallyQA tests the response-format comparison at answer onset.
Fixed-prefix replay, stage-aligned knockout, and KV swaps on synthetic scenes further examine the roles of access history and generated content.
\autoref{fig:conditions} maps the four comparisons reported in \autoref{sec:results}: response format, description type, question-time access, and attribute expression.

% P4 -- the relations found, ordered around the before-expression evidence (review P1, 2026-09-26)
The response-format comparison shows that direct image access can continue to support accuracy after question processing.
Direct answers are robust to answer-side knockout when question-time image access is available, whereas CoT responses lose accuracy.
In an immediate-answer control, the CoT instruction alone does not reproduce this loss.
Supplying the required attributes in text makes generation robust to knockout, whereas a correct history that omits them or a description with the attributes withheld does not.
On synthetic follow-up counting, direct image access remains consequential before attribute expression even when image-conditioned question and prefix states are retained.
Knockout placed before the model writes the required attributes reduces accuracy, and replacing the image KV states at that point shifts answers toward the donor's count.
Both interventions have smaller effects after sufficient expression.
Even when the initial generated prefix is held fixed, later knockout sensitivity still differs between question-time access conditions.
The visible text and the preceding image access therefore both matter.

% P5 -- the takeaway and the controls (review P1, 2026-09-26)
% The main experiments use Qwen2.5-VL-32B \citep{bai2025qwen25vl, qwen2025release}, and several patterns also hold in InternVL3-14B \citep{zhu2025internvl3}.
% The additional comparisons delineate the scope of these patterns across models and tasks.
The main experiments use Qwen2.5-VL-32B \citep{bai2025qwen25vl, qwen2025release} on synthetic and real-image tasks.
Additional experiments with InternVL3-14B \citep{zhu2025internvl3} support several patterns on synthetic tasks and also show differences across tasks and generation stages.
Together, these findings relate image dependence to the information needed for the current question and the history of image access, beyond generation position alone.
This perspective informs decisions on when to reduce visual access during an answer and which information to retain for subsequent questions.

\section{Experimental Framework}
\label{sec:method}

An autoregressive VLM encodes the image as \emph{image tokens}, processes the \emph{question tokens} together with any dialogue history and attribute description, and generates the answer token by token.
We distinguish \emph{question processing} from \emph{answer generation}.
\emph{Direct image access} is attention from question or answer positions to image tokens.
We operationalize \emph{image dependence} as \emph{knockout sensitivity}, defined as the accuracy change caused by blocking specified attention edges to image tokens, relative to a baseline with the same input and question-access condition.
This measure reflects sensitivity to removing direct access rather than how often the model accesses the image during unperturbed generation.
The image-KV swap in \autoref{sec:expression} separately measures how answers respond to changes in image-side content.
\emph{Intervention sensitivity} refers to the effect of either intervention.

\subsection{Attention Knockout}
\label{sec:knockout}

We use attention knockout to block selected attention edges \citep{geva2023dissecting}, following its application to visual information flow in VLMs \citep{kaduri2025whats}.
The mask adds $-\infty$ to the pre-softmax logits from the selected positions to image tokens in every head and renormalizes the remaining weights; previously computed key--value states and text-to-text attention are retained (\appref{app:knockout_impl}).

Question-side and answer-side knockout are controlled independently.
Question-time image access is \emph{Allowed} when the rows of the current question can attend to image tokens and \emph{Blocked} when these edges are masked at every layer.
A preceding turn, image included, is encoded with full access and cached, so Blocked preserves the image information already in those states.
On the answer side, knockout is active at layer $k$ and generation step $t$ when $k\ge\ell$ or $t\ge\tau$, where $\tau$ counts the answer tokens generated before blocking and $\ell=L$ imposes no layer restriction.
Because a layer cutoff $\ell<L$ acts from the first answer token, statements about late generation rest on $\ell=L$.

\subsection{Knockout Loss}
\label{sec:loss}

Let $c$ denote an experimental condition (task, model, inputs, response format, and question-time access), $\Abase(c)$ its accuracy without answer-side knockout, and $\Ako(c;\ell,\tau)$ its accuracy with knockout at $(\ell,\tau)$.
The knockout loss, in percentage points (pp), is
\begin{equation}
\dko(c;\ell,\tau)=\Abase(c)-\Ako(c;\ell,\tau).
\end{equation}
When question-time access is Blocked, the baseline blocks it too, so the loss isolates what answer-side knockout adds.
For the question-access comparison, we define $D=\dko(\mathrm{Blocked})-\dko(\mathrm{Allowed})$ with task, model, history, and cutoff fixed.
A positive $D$ indicates a larger knockout loss under Blocked access.
We interpret this contrast alongside each arm's baseline and knockout accuracy.
A positive loss shows that direct image access contributed to accuracy under that condition, and a negative loss that severing it helped on those examples.
A zero loss shows that severing this pathway at this position did not change aggregate accuracy on the evaluated examples.
Because individual answers can still change in either direction and other pathways may compensate \citep{salazar2026pathways}, a zero loss does not show that the image information is absent or unused.

We report pointwise 95\% confidence intervals (CIs) from a percentile bootstrap over images with pairing preserved \citep{efron1982jackknife}.
Every contrast is computed within one run because greedy CoT outputs can differ between runs.
These intervals quantify image-sampling uncertainty conditional on the run, not variability across runs.
Image sets, their overlap, execution environments, and pairing units are documented in \appref{app:tasks}.
Conclusions rest on cutoffs at which knockout is applied to every response.
In the figures, bands and error bars show the CIs, and open markers indicate cutoffs where knockout is applied to only a subset of responses.

\subsection{Tasks and Information Conditions}
\label{sec:tasks}

\paragraph{Synthetic tasks.}
Synthetic scenes come from a seeded generator, so every object's color, shape, and location is known (\appref{app:tasks}).
As in CLEVR \citep{johnson2017clevr}, the scenes let us control the attributes an answer requires and whether they are available in text.
Single-turn counting asks for the number of objects of a given color among 6--7 objects, with five answer options.
Follow-up counting first asks which quadrant contains a unique anchor object, then asks how many objects of a specified color that quadrant contains.
The first reply is teacher-forced to control the dialogue history.
The \emph{required attributes} are the colors the current question needs, and text that supplies them is \emph{answer-sufficient}.
Description conditions vary the information supplied alongside the question and are \emph{None}, \emph{Complete} (every object's attributes, without the count), \emph{Partial} (the same listing with the colors withheld), or \emph{Unrelated} (token-matched text about something else).
History conditions vary the information in the preceding reply and are \emph{Label-only} (the correct quadrant), \emph{Attribute-stating} (the quadrant and its objects by color and shape), or \emph{Shape-only} (the same at equal length without the colors).
For generated prefixes, \emph{Sufficient} denotes the annotation category defined in \appref{app:annotation}.
\autoref{tab:app_a4} gives the full matrix.

\paragraph{Real-image tasks.}
GQA tests the description and question-time access comparisons on real images.
From GQA \citep{hudson2019gqa}, whose images come from Visual Genome \citep{krishna2017visualgenome}, we select relation-color questions that identify an object through a relation and ask for its color.
We convert each question to a five-choice format for our evaluation.
Complete descriptions are scene-graph listings that include the queried color and serve as an answer-revealing positive control.
TallyQA tests the response-format comparison.
We use its simple questions \citep{acharya2019tallyqa}, which ask how many instances of a noun are present, and balance the selected answers across 1--4.
Both tasks introduce the image in a preceding turn so that question-time access can be controlled as in the follow-up task (\appref{app:tasks}).

\section{What Changes Image Dependence?}
\label{sec:results}

Unless otherwise stated, results use Qwen2.5-VL-32B-Instruct with greedy decoding, and \appref{app:internvl} reports the corresponding analyses for InternVL3-14B; scoring is described in \appref{app:tasks}.
The four comparisons in \autoref{fig:conditions} examine how image dependence varies with response format, description type, question-time image access, and attribute expression during generation.

\subsection{Response Format Changes Image Dependence}
\label{sec:format}

\begin{figure}[t]
\centering
\includegraphics[width=\linewidth]{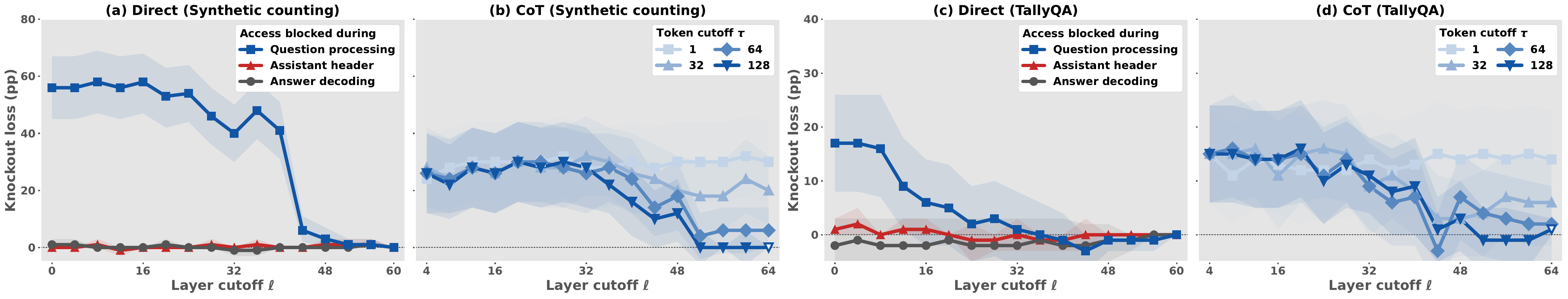}
\caption{\textbf{Comparison of knockout loss across response formats and tasks.} Direct answers are sensitive to question-side knockout in earlier layers, whereas CoT answers also show answer-side loss. \textbf{(a, b)} Synthetic counting; \textbf{(c, d)} TallyQA. Direct panels separately block image access from the question, assistant-header, or decode rows at layers $k\ge\ell$. CoT panels combine answer-side layer and token cutoffs ($k\ge\ell$ or $t\ge\tau$), with question-time access Allowed.}
\label{fig:format_synthetic}
\end{figure}

We first test whether image access during question processing is sufficient for subsequent answer generation across response formats.
\emph{Direct} answers give only the final answer, whereas \emph{CoT} answers follow a zero-shot instruction to reason step by step \citep{kojima2022zeroshot} in at most 120 words before giving the answer.
An \emph{immediate-answer control} keeps the CoT instruction verbatim but opens the answer at once; it shows no observed loss on either task (\appref{app:format_controls}), so the instruction alone does not explain the CoT sensitivity.
In the matched answer-onset controls in \appref{app:format_controls}, all three formats use the same images and answer-side knockout within each task.

On synthetic counting, \autoref{fig:format_synthetic}(a) shows the knockout loss of Direct answers when image access is blocked separately from the question, assistant-header, or decode rows.
Blocking question rows at layers $k\ge\ell$ reduces accuracy for cutoffs below about 44, with no observed loss at higher cutoffs.
Blocking assistant-header or decode rows produces no observed loss at any tested layer cutoff.
Thus, Direct answers depend on image access during question processing but are robust to answer-side knockout when question-time access is Allowed.
CoT answers remain sensitive during generation.
With no layer restriction, the loss at $\tau=32$ is about two thirds of that at $\tau=1$ and becomes small at $\tau=64$ (\autoref{fig:format_synthetic}(b)).
Layer-only knockout likewise produces losses comparable to blocking all layers when the cutoff is $\ell\le40$.
The same ordering holds on an independent count-balanced set of 100 images with decision rules fixed before the run (\appref{app:format_controls}).
The Direct profile is consistent with the answer being formed from states computed during question processing \citep{salazar2026pathways}, although it does not show where the answer is stored.

The same response-format contrast holds on real images in TallyQA (\autoref{fig:format_synthetic}(c, d)).
For Direct answers, blocking question rows reduces accuracy at layer cutoffs below about 24, while blocking header or decode rows produces no observed loss.
CoT answers lose accuracy under answer-side knockout.
The CoT loss has a layer profile similar to that on synthetic counting, but with no layer restriction it fades earlier, by $\tau=32$.
Without knockout, CoT is not more accurate than Direct on this task, consistent with reports that CoT does not always improve visual reasoning \citep{jin2026look, kancheti2026cotdegrades}.
Under knockout, CoT accuracy falls below the Direct baseline on TallyQA, but only to that baseline on the fresh synthetic set (\appref{app:format_controls}).
Thus, direct image access can support CoT accuracy even when CoT offers no advantage over Direct.

Across synthetic counting and TallyQA, direct image access continues to support accuracy during CoT generation, while Direct responses are robust to answer-side knockout.
The comparison therefore establishes a difference between generation protocols but does not isolate reasoning itself, since the protocols also differ in generated content, length, and computation \citep{pfau2024dotbydot}.
We next examine whether supplying the required visual attributes in text reduces this dependence.

\subsection{Description Type Changes Image Dependence}
\label{sec:description}

We test whether supplying the attributes needed for the current question reduces dependence on direct image access, and whether a correct prior answer provides the same robustness.
We compare the four description types in single-turn and follow-up counting and the three history types in follow-up counting, with question-time access Allowed unless stated.

\begin{figure}[t]
\centering
\includegraphics[width=\linewidth]{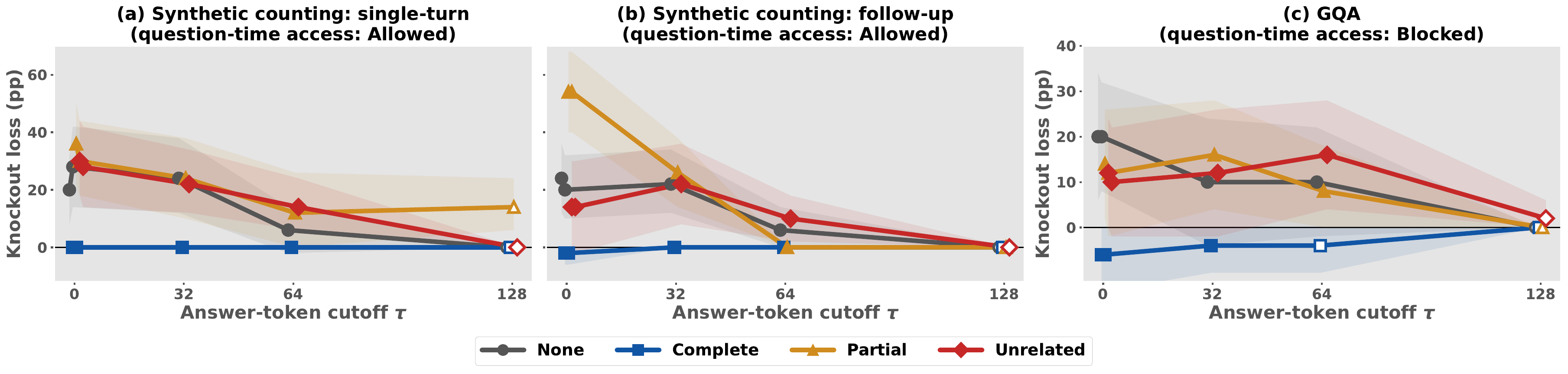}
\caption{\textbf{Comparison of knockout loss across description conditions.} Complete descriptions yield little observed loss across the measured cutoffs; descriptions omitting the required attributes do not consistently provide the same robustness. \textbf{(a)} Single-turn synthetic counting and \textbf{(b)} synthetic follow-up counting with the Label-only history, both with question-time access Allowed. \textbf{(c)} GQA with question-time access Blocked. All panels use CoT and token cutoffs without a layer restriction.}
\label{fig:desc_synthetic}
\end{figure}

As \autoref{fig:desc_synthetic}(a) shows for single-turn counting, supplying the required colors in a Complete description leaves aggregate accuracy unchanged under answer-side knockout at every measured cutoff.
Without a description, the loss appears from answer onset and persists across later cutoffs.
Complete serves as a positive control for answering from supplied visual attributes, as in systems that respond to visual questions from captions or tags \citep{yang2022pica, zeng2022socratic}.
Partial descriptions withhold the colors, while Unrelated descriptions provide token-matched text without the required information.
Neither reproduces the robustness of Complete, and both show losses comparable to or larger than those under None through $\tau=64$.
Thus, the content of the description, beyond its added length, supports robustness in this setting.

Complete descriptions also remove the loss in follow-up counting, where the correct Label-only history omits the colors needed for the current question (\autoref{fig:desc_synthetic}(b)).
Under Partial descriptions, the loss at early cutoffs exceeds that of None, from a higher baseline, and vanishes by $\tau=64$.
Unrelated text yields smaller loss estimates than None at the first two cutoffs and the same loss at $\tau=32$, with overlapping intervals.
Thus, Complete provides robustness in both synthetic settings, whereas text without the required attributes can change the size of the loss without removing it.

The description comparison extends to real images in GQA relation-color questions (\autoref{fig:desc_synthetic}(c)).
With question-time access Blocked, answer-side knockout reduces accuracy from answer onset.
With access Allowed, the observed loss is small at every measured cutoff (\appref{app:desc_history}).
Complete descriptions remove the Blocked loss, whereas color-withholding Partial descriptions and unrelated text of matched length do not (\autoref{fig:desc_synthetic}(c)).
Because Complete includes the queried color, this result establishes robustness when answer-relevant information is supplied in text.

\begin{figure}[t]
\newsavebox{\figfivebox}\newsavebox{\tabonebox}
\sbox{\figfivebox}{\includegraphics[width=0.47\linewidth]{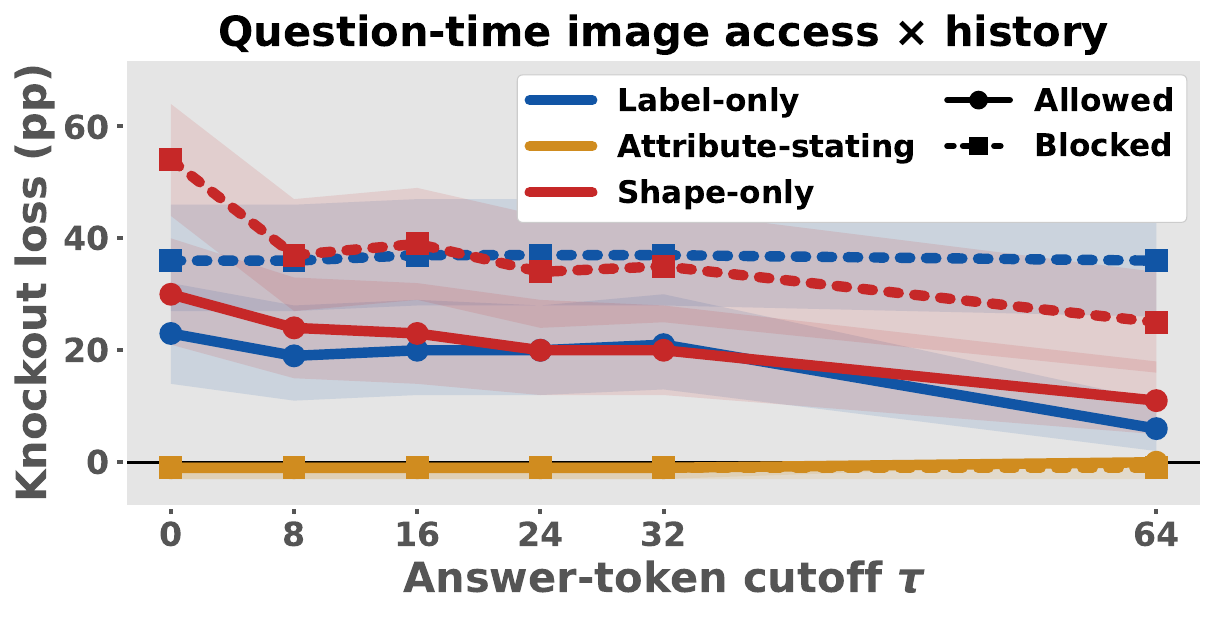}}
\sbox{\tabonebox}{{\scriptsize\setlength{\tabcolsep}{2.5pt}\renewcommand{\arraystretch}{1.1}
\begin{tabular}{@{}l l c r@{}}
\toprule
\textbf{Prefix source} & \textbf{Replay access} & \textbf{Accuracy (\%)} & \textbf{Loss (pp)} \\
\midrule
Allowed & Allowed & 87 $\to$ 81 & 6 [2, 11] \\
 & Blocked & 80 $\to$ 66 & 14 [8, 21] \\
\rowcolor{blue!8} & \multicolumn{2}{l}{\textbf{Difference: Blocked $-$ Allowed}} & \textbf{8 [1, 15]} \\
\midrule
Blocked & Allowed & 83 $\to$ 67 & 16 [8, 25] \\
 & Blocked & 69 $\to$ 33 & 36 [26, 46] \\
\rowcolor{blue!8} & \multicolumn{2}{l}{\textbf{Difference: Blocked $-$ Allowed}} & \textbf{20 [9, 31]} \\
\bottomrule
\end{tabular}}}
\newlength{\topboxht}\setlength{\topboxht}{\maxof{\ht\figfivebox+\dp\figfivebox}{\ht\tabonebox+\dp\tabonebox}}
\begin{minipage}[t]{0.47\linewidth}\vspace{0pt}
\centering
\parbox[c][\topboxht][c]{\linewidth}{\centering\usebox{\figfivebox}}
\caption{\textbf{Knockout loss across history and question-time access.}
Attribute-stating histories yield near-zero knockout loss under both access conditions.
Follow-up counting, CoT, independent set; each condition uses its own baseline.}
\label{fig:qaccess}
\end{minipage}\hfill
\begin{minipage}[t]{0.50\linewidth}\vspace{0pt}
\centering
\parbox[c][\topboxht][c]{\linewidth}{\centering\usebox{\tabonebox}}
\captionof{table}{\textbf{Fixed-prefix accuracy and knockout loss.} The same 64-token prefix is replayed under Allowed and Blocked with states recomputed; both prefix sources retain a positive loss difference. Accuracy is baseline $\to$ knockout at $\tau=64$; CoT, Label-only history.}
\label{tab:prefix}
\end{minipage}
\end{figure}

We then ask whether the same distinction holds for information supplied in the preceding reply.
The history comparison in \autoref{fig:qaccess} uses an independent 100-image set and separates correctness on the preceding question from sufficiency for the current one (InternVL3 in \appref{app:internvl}).
The Label-only history is correct, but the follow-up needs colors that it never states.
The Attribute-stating history states them, and the Shape-only history has the same token count and structure without them.
With the Attribute-stating history, knockout loss remains near zero in both question-time access conditions across the measured cutoffs, and the Blocked baseline rises to the level of the Allowed one.
The Shape-only history does not reproduce this tolerance.
Both arms lose accuracy under knockout, and the question-access difference remains.
The Shape-only losses exceed the Label-only ones, but so do its baselines, so we read the result as a failure to reproduce the tolerance at matched length, not as harm from the added text.

In the main model, supplying the required attributes supports robustness to answer-side knockout across synthetic tasks and GQA.
A correct prior answer or text that omits those attributes does not consistently provide the same robustness.
InternVL3-14B shows the same distinction in single-turn counting and the history comparison, while its follow-up description conditions separate less clearly (\appref{app:internvl}).
These comparisons identify the relevance of the supplied information to the current question as a key factor in image dependence.
We next hold the history fixed and examine how image access during question processing affects subsequent dependence.

\subsection{Question-Time Access Changes Image Dependence}
\label{sec:qaccess}

When the required attributes are not in the text, does image access while the question is processed change the later sensitivity to answer-side knockout?
We hold the history fixed (Label-only, correct) and vary only question-time image access to test its effect on subsequent answer-side knockout sensitivity.
Answer-side knockout is then applied at each $\tau$, and each arm is compared with its own baseline, so the cost of blocking question-time access itself is separated from the later loss.

\paragraph{Free generation.}
With the Label-only history, answer-side knockout causes a larger loss under Blocked than Allowed at every measured cutoff on the independent synthetic follow-up set (\autoref{fig:qaccess}).
The sustained Blocked loss accompanies a specific failure pattern: knocked-out responses answer ``0'' at early cutoffs on the profile set and mostly state that only the Turn~1 object is present on the independent set (\appref{app:qaccess_prefix}).
The sustained Blocked loss therefore reflects a recurring failure of the continuation, rather than a graded measure of information retained during question processing.

\paragraph{Fixed-prefix replay.}
The access conditions also change the generated text: at the replay cutoff, some Allowed responses already state the answer-sufficient colors, whereas none of the Blocked responses do (\autoref{fig:app_c5}).
The smaller Allowed loss may therefore reflect differences in the visible prefix.
To test whether this explains the contrast, we replay the same 64-token prefix under both access conditions, recompute the states, and apply knockout from token 65 (\appref{app:replay}).
For either prefix source, knockout loss remains larger under Blocked than Allowed access (Table~\ref{tab:prefix}).
With an Allowed-derived prefix, this difference persists while both knockout accuracies remain above the constant-answer reference; the Blocked-derived comparison includes a cell at that reference.
Thus, differences in the initial visible text do not fully explain the question-access effect.
The two prefix sources define separate comparisons, rather than a decomposition of the free-generation difference.
Because prefix states are recomputed and subsequent text is generated freely, the experiment does not identify which retained states account for the difference.

In Qwen2.5-VL-32B, preceding image access affects subsequent knockout sensitivity even with the same visible prefix.
In free generation, InternVL3-14B shows a question-access difference at answer onset but not at later cutoffs (\appref{app:internvl}); the fixed-prefix result has not been established for this model.

\subsection{Image Dependence Changes with Attribute Expression}
\label{sec:expression}

The description comparison supplied the required attributes from outside.
Does sensitivity also change when they appear in the model's own text during generation?
A first indication does not depend on stage cutoffs: at token 64, Sufficient prefixes in the Allowed arm show no observed knockout loss, whereas Uninformative prefixes show loss in both access arms (\autoref{fig:app_c5}).

\paragraph{Stage-aligned cutoffs.}
A fixed token cutoff falls at different stages of different responses.
We therefore give each response its own cutoff, determined by a classifier whose rule was fixed after inspecting the 64-token prefixes with answer correctness hidden and before any cutoff run (\appref{app:annotation}).
\emph{Before expression} places knockout at the prediction of the token at which the first non-anchor object color is written.
\emph{After expression} places it at the token after the prefix is first labeled Sufficient, meaning that the prefix states both required colors or gives a closed enumeration that implies the correct count.
The two positions are content-defined stages separated by several descriptive steps, not the moments around one event; their absolute positions differ between arms, and a response enters a stage only if it reaches that position, so the stages cover different response subsets.

\begin{wrapfigure}{r}{0.4\linewidth}
\vspace{-10pt}
\centering
\includegraphics[width=\linewidth]{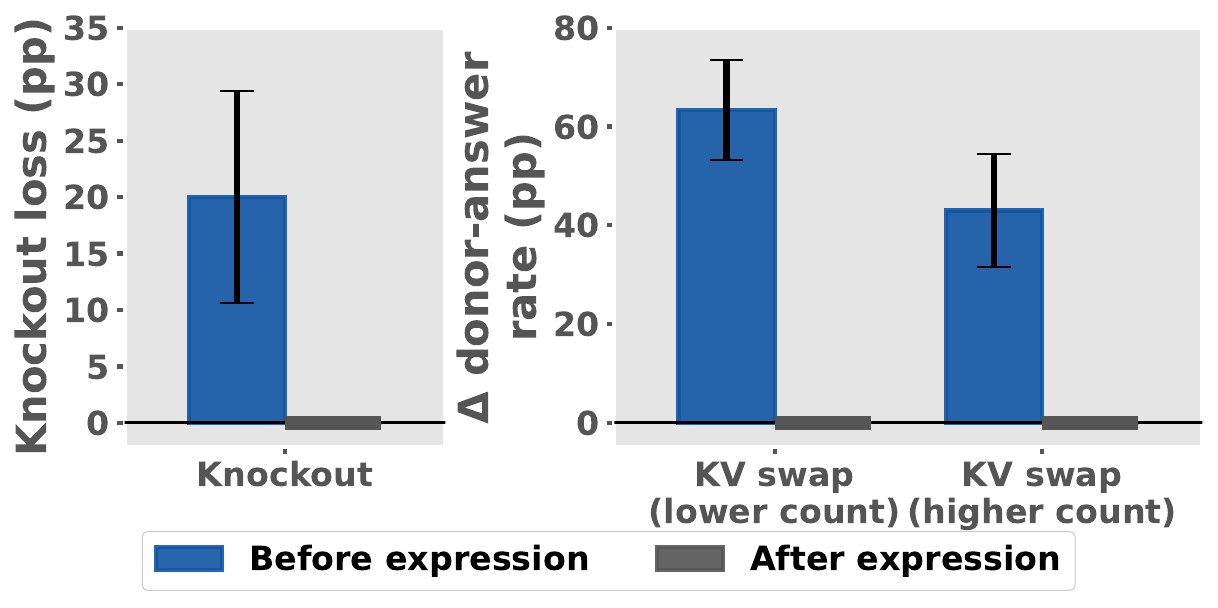}
\caption{\textbf{Intervention effects before and after expression.} Both effects are smaller after expression. \textbf{Left:} knockout loss on the images reaching both stages. \textbf{Right:} change in donor-answer rate relative to no swap, for the donors with the smaller and larger alternative counts. Question-time access Allowed.}
\label{fig:expression}
\vspace{-8pt}
\end{wrapfigure}

On the same images reaching both stages in the Allowed arm, knockout reduces accuracy before expression but produces no observed aggregate loss after expression (\autoref{fig:expression}, left).
The Blocked arm and the all-eligible populations show the same qualitative pattern (\appref{app:stage_more}).
Thus, direct access remains consequential before expression even though previously computed image-conditioned states are retained.
The rule uses the ground-truth colors, an oracle annotation for analysis rather than a deployment signal.
It was developed on the 64-token prefixes of this same image set and has no independent annotator, and the uncertainty of the cutoff positions is not represented in the intervals (\appref{app:annotation}).

\paragraph{Counterfactual image-KV swap.}
Knockout alone leaves open whether the before-expression loss reflects sensitivity to visual content or disruption caused by blocking access.
We therefore replace the image-position KV states in every layer with those from a donor image at each stage, while retaining the recipient's text-position states.
In the donor image, the target objects are recolored so that the correct count changes to one of the two alternative values.
Generation continues from the recipient's prefix, a directional form of activation patching \citep{zhang2024activationpatching, heimersheim2024patching} (\appref{app:swap}).

We measure the change in donor-answer rate relative to the no-swap continuation in the same run.
Self swaps reproduce that continuation token for token on every checked item, while recoloring an object outside the target quadrant shows no clear accuracy loss (\appref{app:swap_controls}).
On the same images across stages, both donors increase donor-answer rates before expression but produce no observed increase after expression (\autoref{fig:expression}, right).
These directional changes show that answers are sensitive to the inserted visual content before expression, arguing against a uniform-disruption account.

Together, the interventions show sensitivity to direct image access and image-side content before expression, with smaller effects after expression.
This complements the supplied-description results: reduced sensitivity is also observed after the required information appears in the model's own text.
The comparison does not locate the transition or establish whether verbalization itself, accumulated states, or answer commitment causes the change.
These stage-aligned results are established on synthetic follow-up counting with Qwen2.5-VL-32B.

\section{Related Work and Discussion}
\label{sec:discussion}

\paragraph{From visual pathways to dependence during generation.}
Three findings frame our question.
Visual information can be transferred into question-token representations \citep{kaduri2025whats, zhang2025crossmodal}, and under some interventions models fall back on a text-mediated pathway when their usual pathway is ablated \citep{salazar2026pathways}.
Across dialogue turns, facts stated in an earlier reply can support a later answer from assistant-text KV states, whereas unstated facts leave the answer dependent on the visual states \citep{chen2026seen}.
In long generations, the contribution of image tokens concentrates early in the response \citep{sammani2026scaling}.
Our study connects these findings to how information needed for the current question becomes available during answer generation.
% The same study removes image KV entries during generation at fixed token cutoffs and measures the resulting accuracy \citep{sammani2026scaling}.
\citet{sammani2026scaling} also remove image KV entries during generation at fixed token cutoffs and measure the resulting accuracy.
Other nearby designs change the image \citep{shi2026seeingsaying}, change question presentation \citep{ask2026twice}, re-complete a truncated reasoning chain without the image \citep{sun2025tvc}, or compare stated and unstated facts across dialogue turns \citep{chen2026seen}.
We add interventions aligned to attribute expression, directional image-KV replacement, and a comparison of question-time access under fixed visible prefixes.

\paragraph{Image dependence beyond visible text.}
The description and history comparisons show that supplying attributes needed for the current question supports robustness to knockout, whereas a correct prior answer that omits them does not.
The stage comparisons extend this observation to the model's own generation: sensitivity to both knockout and image-KV replacement is smaller after expression.
Fixed-prefix replay further shows that the same visible text can accompany different knockout sensitivity under different question-time access conditions.
Together, these results relate image dependence to both the information expressed in context and the access history under which that context was computed.

\paragraph{Interpreting reduced dependence after expression.}
One explanation is that verbalizing relevant attributes makes visual evidence reusable for subsequent computation, as written intermediate steps can do in language models \citep{nye2021scratchpads}.
This account is consistent with reports that VLMs rely on textual anchors and under-use available visual representations \citep{shahgir2026vlmsneedwords, fu2025hiddenplainsight, liu2025visualrepresentations}.
However, attribute expression also accompanies changes in retained states and may follow an already committed answer; stated reasoning need not reveal what determined the answer \citep{turpin2023unfaithful}.
Our results establish a difference in intervention sensitivity across the two stages, while leaving these mechanisms unresolved.
Interventions on generated prefixes may help test the contribution of stated content \citep{lanham2023faithfulness}, with matched re-encoding controls to account for changes in prefix states.

\paragraph{Sufficiency is question-dependent.}
The correct quadrant label supports the first answer but omits the colors needed for the follow-up question.
This illustrates two distinct decisions: whether direct image access can be reduced for the current answer, and which visual information should remain available for later questions.
% Our findings motivate access policies that consider the information needed by the current question and its availability in context, while retaining access to information that the dialogue has not expressed \citep{chen2026seen}.
Our findings motivate access policies that consider the information needed by the current question and its availability in context.
Reduced dependence for the current answer may justify restricting visual access temporarily, but it does not imply that the underlying visual information can be discarded permanently.
Because a subsequent question may require attributes absent from the existing text \citep{chen2026seen}, such policies also need to preserve or restore access to the image.
Methods that select or prune visual tokens \citep{chen2024fastv, yang2025visionzip, zhang2025sparsevlm, khaki2025sparsevila}, freeze visual-token updates in selected layers \citep{yuan2025shortv}, or compress and evict visual KV states \citep{tu2025vlcache, wan2024lookm, li2025madakv, huang2025aircache, liu2026retentivekv} face these decisions when reducing visual computation or memory.
Other methods re-inject or re-attend to the image during generation \citep{sun2025tvc, chu2025lookagain, zou2025memvr}, or explicitly control visual evidence use through dynamically triggered evidence insertion \citep{li2026aimcot}, alternating looking and thinking steps coupled with compression \citep{cai2026lookandthink}, or perception tokens that the model generates to trigger additional perception \citep{yu2025perceptiontokens}.
Our comparisons offer diagnostic conditions for evaluating whether such strategies preserve access to question-relevant information.
They do not yet provide an online sufficiency detector or establish an accuracy--efficiency trade-off.

\paragraph{Scope and limitations.}
The response-format and information-condition comparisons provide evidence across synthetic and real-image tasks, with several patterns also observed in InternVL3-14B (\appref{app:internvl}).
The strength and timing of these patterns vary across models and tasks.
The stage-aligned knockout and image-KV swap results are established only for Qwen2.5-VL-32B on synthetic follow-up counting.
Common-set analyses align the images compared across stages but do not control for elapsed generation or difficulty (\appref{app:stage_more}, \appref{app:swap_controls}).
Finally, these interventions measure sensitivity under altered access conditions: preserved states, compensatory pathways \citep{salazar2026pathways}, and the out-of-distribution question-side mask limit what they imply about image use during unperturbed generation.

\section{Conclusion}
\label{sec:conclusion}

% Image dependence during answer generation varies with how the information needed for the current question becomes available in context. The response-format comparison shows that direct image access can continue to support accuracy during CoT generation, whereas direct answers are robust to its removal. Across real-image and synthetic tasks, supplying the required attributes in text reduces sensitivity to blocking direct image access. On synthetic follow-up counting, sensitivity to both knockout and image-KV replacement is smaller after attribute expression, while fixed-prefix replay shows that preceding image access matters even when the visible prefix is identical. Together, these findings motivate considering both question-relevant information and access history when deciding when to reduce visual access during an answer and which visual information to retain for subsequent questions.
Image dependence during answer generation varies with how the information needed for the current question becomes available in context, rather than with generation position alone.
The response-format comparison shows that direct image access can continue to support accuracy during CoT generation, whereas direct answers are robust to answer-side knockout when question-time access is available.
Across real-image and synthetic tasks, supplying the required attributes in text reduces this dependence.
On synthetic follow-up counting, direct image access remains consequential before the model states the required attributes, even though image-conditioned question and prefix states are retained.
At this stage, knockout reduces accuracy and image-KV swaps move answers toward the donor's count; both effects are smaller once the attributes are stated.
Fixed-prefix replay further shows that the same visible prefix can accompany different sensitivity depending on the preceding image access.
Retaining image-conditioned states therefore does not by itself make direct image access dispensable.
 These findings motivate considering both question-relevant information and access history when deciding when to reduce visual access during an answer and which visual information to retain for later questions.

\label{maintextend}
\clearpage
\section*{Reproducibility statement}
Estimates use saved per-image outcomes, paired input identities, and condition-matched baselines; source records retain image and output hashes, settings, annotations, and intervention diagnostics.
\appref{app:method} specifies the knockout, generation, scoring, replay, and KV-swap protocols.
\section*{Ethics statement}
This study uses synthetic scenes and benchmark images.
Accurate attributes are experimental controls, not an assumption that reliable descriptions are available in deployment.
Results do not establish safe access removal outside the evaluated conditions.
\section*{Statement on the use of large language models}
In this work, we used generative AI tools to propose and refine hypotheses, to design and provide feedback on the research methodology and experiments, to implement methods (intervention code for attention knockout, image-KV swap, and fixed-prefix replay, and the analysis pipelines), to develop the rule that labels attribute expression in generated prefixes, to assist with translation of drafts written in Japanese, and to interpret results.
We have not used generative AI tools to generate the synthetic data sets: all scenes are produced by a deterministic, seeded generator, whose code was written with AI assistance.
Developing theoretical models, formulating mathematical claims, and writing proofs are not applicable to this work.
Additionally, we used generative AI tools to create and edit software code, create and modify figures, draft parts of the paper, edit the paper for readability, identify related literature and check bibliographic metadata, format references, and suggest the structure and title of the paper.
We have reviewed all AI-assisted work: the intervention and analysis code was checked with parity tests and against saved per-image outcomes (\appref{app:swap}), the attribute-expression rule was frozen before any stage intervention ran and its limits are reported (\appref{app:annotation}), every cited work was checked against its source by the authors, and all numbers in the text were checked against the saved results.
We take responsibility for the final content of this work, including text, claims, and artifacts produced with the aid of generative AI.

\bibliography{references}
\bibliographystyle{iclr2027_conference}

\clearpage
\section*{Appendix Contents}
\begingroup\small
\newcommand{\appindexline}[2]{\noindent\hyperref[#1]{#2}\hfill\pageref{#1}\par}
\appindexline{app:method}{\textbf{A. Experimental Details}}
\appindexline{app:knockout_impl}{\quad A.1 Attention Knockout Implementation}
\appindexline{app:loss_uncertainty}{\quad A.2 Metrics and Uncertainty}
\appindexline{app:tasks}{\quad A.3 Evaluation Data and Information Conditions}
\appindexline{app:prompts}{\quad A.4 Prompts}
\appindexline{app:annotation}{\quad A.5 Attribute-Expression Annotation and Intervention Stages}
\appindexline{app:replay}{\quad A.6 Fixed-Prefix Replay}
\appindexline{app:swap}{\quad A.7 Counterfactual Image-KV Swap and Implementation Checks}
\medskip
\appindexline{app:models}{\textbf{B. Model Family Ablations (InternVL3-14B)}}
\appindexline{app:internvl_format}{\quad B.1 Response Format}
\appindexline{app:internvl_description}{\quad B.2 Description Type}
\appindexline{app:internvl_history}{\quad B.3 History Content and Question-Time Access}
\medskip
\appindexline{app:controls}{\textbf{C. Supplementary Controls}}
\appindexline{app:format_controls}{\quad C.1 Immediate-Answer Control and Independent Confirmation}
\appindexline{app:desc_history}{\quad C.2 Additional Information Conditions}
\appindexline{app:qaccess_prefix}{\quad C.3 The Conditions behind the Fixed-Prefix Replay}
\appindexline{app:stage_more}{\quad C.4 Attribute Stages and Common Sets}
\appindexline{app:swap_controls}{\quad C.5 Counterfactual Swap Controls}
\endgroup

\clearpage
\appendix
\raggedbottom
\counterwithin{table}{section}
\counterwithin{figure}{section}
\renewcommand{\topfraction}{0.95}\renewcommand{\bottomfraction}{0.9}\renewcommand{\textfraction}{0.03}\renewcommand{\floatpagefraction}{0.85}
\makeatletter\setlength{\@fptop}{0pt}\setlength{\@fpsep}{10pt plus 2pt}\setlength{\@fpbot}{0pt plus 1fil}\makeatother
% Appendix policy (2026-09-26): completed comparisons and the specifications / limits needed to evaluate them only.
% Experiment plans, re-aggregations still pending, and production to-dos live in docs/appendix_todo.md.

% ==== Appendix rewritten 2026-09-26 from the user's Japanese draft (appendix restructure). The previous appendix is kept below inside \iffalse ... \fi.
\providecommand{\figplaceholder}[2][3.6cm]{\fbox{\parbox[c][#1][c]{0.94\linewidth}{\centering\footnotesize\textcolor{red!70!black}{\textbf{[Figure to be added]}}\\[3pt]\parbox{0.9\linewidth}{\centering #2}}}}

\section{Experimental Details}
\label{app:method}

\subsection{Attention Knockout Implementation}
\label{app:knockout_impl}

\autoref{tab:app_a1} lists the attention edges severed by the interventions of \autoref{sec:knockout} and by the comparisons of \autoref{fig:format_synthetic}.
Every intervention adds $-\infty$ to the pre-softmax logits, in every head, from the selected query rows to the image-token key columns, and the remaining attention weights are renormalized.
Image tokens themselves, the attention edges that are not targeted, and the key--value (KV) states computed before the intervention are retained.

\begin{table}[h]
\centering
\footnotesize
\setlength{\tabcolsep}{3pt}
\renewcommand{\arraystretch}{1.15}
\caption{\textbf{Targets and onset of attention knockout.}
Decoder layers are indexed $k=0,\dots,L-1$.
Ordinary answer-side knockout severs the edges whenever either the layer condition or the token condition holds; the row-group sweeps are a separate comparison that severs one kind of row only.}
\label{tab:app_a1}
\begin{tabularx}{\linewidth}{@{}>{\raggedright\arraybackslash}p{2.2cm} >{\raggedright\arraybackslash}p{2.6cm} >{\raggedright\arraybackslash}p{2.1cm} >{\raggedright\arraybackslash}X >{\raggedright\arraybackslash}p{3.0cm}@{}}
\toprule
\textbf{Intervention} & \textbf{Query rows} & \textbf{Layer condition} & \textbf{Where it applies} & \textbf{Retained computation} \\
\midrule
Question-side knockout & rows of the current user question & all layers & while the current question is processed & KV of earlier turns; edges from the assistant-header and decode rows \\
\hl{Answer-side knockout} & assistant-header rows and the decode rows that follow & $k\ge\ell$, or all layers once the token condition holds & from the position set by the token cutoff $\tau$; the layer condition also acts before it & KV computed before the intervention; text-to-text attention \\
Question-row sweep & rows of the current user question only & $k\ge\ell$ & while those rows are processed & image access from the header and decode rows \\
Header-row sweep & assistant-header rows only & $k\ge\ell$ & while the header is processed & image access from the question and decode rows \\
Decode-row sweep & decode rows only & $k\ge\ell$ & while the decode rows are processed & image access from the question and header rows \\
\bottomrule
\end{tabularx}
\end{table}

The last assistant-header row predicts answer token~1.
$\tau=0$ severs access from every assistant-header row; $\tau=m>0$ keeps image access for the first $m$ answer tokens and severs it from the prediction of token $m+1$.
$\tau=\infty$ applies no token-based knockout, and $\ell=L$ no layer-based knockout.
Because the two conditions combine by \emph{or}, a layer cutoff $\ell<L$ acts from the first answer token whatever $\tau$ is; statements about late generation therefore use $\ell=L$, and layer effects are read at $\tau=\infty$.

In follow-up and real-image inputs, Turn~1, which contains the image, the first question, and the fixed reply, is computed once with image access, and its KV is kept.
Question-side knockout applies to the rows of the Turn~2 question, and the assistant header is treated as the answer side.
The row-group sweeps run at batch size~1.

Knockout counts as \emph{applied} when, in the forward passes actually executed, at least one answer-side edge to an image token is severed.
When a response ends before the cutoff, the token condition never applies.
Curves with such cutoffs report the application rate, and the aggregate loss, which includes responses without the intervention, is distinguished from the loss over the responses that were actually knocked out.

\clearpage
\subsection{Metrics and Uncertainty}
\label{app:loss_uncertainty}

\autoref{tab:app_a2} summarizes the knockout loss, the question-access difference, and the KV-swap measures of the main text.
Each comparison keeps the pairing by image and uses a baseline from the same execution condition.

\begin{table}[h]
\centering
\footnotesize
\setlength{\tabcolsep}{3.5pt}
\renewcommand{\arraystretch}{1.2}
\caption{\textbf{Measures and the conditions each pairs.}
$A$ denotes accuracy and $p$ the rate of a given answer; rate differences are in percentage points (pp).}
\label{tab:app_a2}
\begin{tabularx}{\linewidth}{@{}>{\raggedright\arraybackslash}p{3.3cm} >{\raggedright\arraybackslash}p{3.6cm} >{\raggedright\arraybackslash}X c@{}}
\toprule
\textbf{Measure} & \textbf{Definition} & \textbf{Held fixed in the pairing} & \textbf{Unit} \\
\midrule
\hl{Knockout loss} $\dko$ & $\Abase-\Ako$ & image, input, model, response format, question-time access & pp \\
Question-access difference $D$ & $\dko^{\mathrm{Blocked}}-\dko^{\mathrm{Allowed}}$ & image, history, cutoff; in replay also the prefix source & pp \\
Donor-answer rate & fraction of answers equal to the donor image's correct count & the specified donor count & \% \\
\hl{Change in donor-answer rate} & $p_{\mathrm{target\ swap}}-p_{\mathrm{no\ swap}}$ & image, prefix, execution path, donor count & pp \\
\bottomrule
\end{tabularx}
\end{table}

When question-time access is Blocked, the baseline also blocks question-side access, so the knockout loss measures the change that answer-side knockout adds to the effect of question-side blocking.
A zero loss means that aggregate accuracy did not change; it does not mean that individual answers are unchanged or that the retained states carry no image information.

The 95\% confidence intervals (CIs) are percentile bootstrap intervals over images \citep{efron1982jackknife}.
Resampling keeps the pairing of the compared conditions, and the several donors of one image are not counted as independent samples.
We draw 20,000 resamples, with seed 20260915 for the September 2026 sweeps and 20260923 for the additional campaigns.
An image missing one condition of a paired contrast is excluded from that contrast.
Intervals on curves are pointwise, not simultaneous over the curve.

When no answer changes in any example, the bootstrap interval degenerates to $[0,0]$, which does not show that the population change rate is exactly zero.
With zero changes in $n$ independent examples, the upper limit of the two-sided 95\% exact binomial interval for the change rate is $1-0.025^{1/n}$.

\clearpage
\subsection{Evaluation Data and Information Conditions}
\label{app:tasks}

\autoref{tab:app_a3} lists the evaluation sets and settings of \autoref{sec:tasks} and \autoref{sec:format}--\ref{sec:qaccess}, and \autoref{tab:app_a4} the information supplied as text.
Conditions compared within one experiment use the same images, and no set is filtered by model correctness.

Runs use batch size~1 with prefill chunk 128, or batch size~4 without chunking.
These settings change some greedy CoT outputs (the single-turn exploration set loses 24~pp at $\tau=0$ in one setting and 28~pp in the other), so the baseline and the intervention of every comparison run in the same setting.
Image-KV swaps run at batch size~1 and are compared with the no-swap continuation of the same run.

Answers are scored on the final \texttt{Answer: $x$} line: an option letter for single-turn and GQA, a nonnegative integer for follow-up and TallyQA.
An answer is correct when exactly one such line exists, it is the last line, $x$ is a valid answer, and $x$ matches the ground truth.
Format violations and responses that reach the output cap count as incorrect and stay in the denominator.

\paragraph{Scene generation and image selection.}
Single-turn scenes place 6--7 non-touching objects on a white $256{\times}256$ canvas; colors are drawn from blue, red, yellow, and green, shapes from star, circle, triangle, and square, radii from 18, 24, 30, and 36~pixels, and angles from $0^\circ$, $45^\circ$, $90^\circ$, and $135^\circ$.
Each question asks for the number of objects of one color.
As in the task design of CLEVR \citep{johnson2017clevr}, controlled attributes and relations make the required information explicit.
Follow-up scenes use $512{\times}512$ images with fixed candidate slots in each quadrant.
The anchor quadrant and one decoy quadrant hold three objects each, the anchor's color--shape pair is unique, and the asked color differs from the anchor's color.
The anchor quadrant holds $c\in\{0,1,2\}$ objects of the asked color and the decoy $3-c$, so the image-wide count never equals the local answer; quadrant and answer are balanced within blocks of 12 items.
GQA questions ask for a color together with a spatial, wearing, or holding relation; yes/no and counting questions are excluded, scene graphs have 2--40 objects, one question per image is chosen by a seeded hash, and the options are the answer color and four distractors from a 15-color list.
We select simple object-count questions from TallyQA \citep{acharya2019tallyqa} on Visual Genome images \citep{krishna2017visualgenome}.
For our evaluation, we retain one question per image and exclude questions about body parts and abstract nouns.

\clearpage
\begin{table}[h]
\centering
\footnotesize
\setlength{\tabcolsep}{3.5pt}
\renewcommand{\arraystretch}{1.12}
\caption{\textbf{Evaluation sets and execution settings.}
Set names in Panel~A denote their roles in separate experiments, and several sets of one task are never pooled into one sample; the GQA 100-image set is a subset of the 300-image set.
No paired contrast crosses the setting differences of Panel~B.}
\label{tab:app_a3}
\textbf{Panel A: evaluation sets}\\[2pt]
\begin{tabularx}{\linewidth}{@{}>{\raggedright\arraybackslash}p{3.1cm} >{\raggedright\arraybackslash}p{3.6cm} r >{\raggedright\arraybackslash}X@{}}
\toprule
\textbf{Set} & \textbf{Task and source} & \textbf{$N$} & \textbf{Selection and answers} \\
\midrule
Single-turn exploration & synthetic generator, validation split & 50 & $256{\times}256$, 6--7 objects; counts 1--5: 14/21/12/2/1 \\
Single-turn fresh & synthetic generator, same style & 100 & counts 1--5, 20 each; generated after exploration, decision rules fixed before the run \\
Single-turn training-style & synthetic generator, training split & 100 & items 0--99; five options (Direct row-group sweep of \autoref{fig:format_synthetic}(a)) \\
Follow-up exploration & synthetic quadrant scenes & 50 & counts 0/1/2: 16/17/17 \\
Follow-up independent & same synthetic generator & 100 & counts 0/1/2: 33/34/33; generated after exploration \\
Follow-up profile & same synthetic generator & 100 & counts 0/1/2 \\
GQA main & GQA relation-color, validation split & 50 & seeded-hash ranks 51--100; five options \\
GQA confirmation & GQA relation-color, training-split candidates & 300 & five options \\
GQA matched subset & subset of the confirmation set & 100 & the same images evaluated on both models \\
TallyQA simple & simple questions on Visual Genome images & 100 & counts 1--4, 25 each \\
\bottomrule
\end{tabularx}

\medskip
\textbf{Panel B: models and execution settings}\\[2pt]
\begin{minipage}{\linewidth}
\renewcommand{\footnoterule}{}
\begin{tabularx}{\linewidth}{@{}>{\raggedright\arraybackslash}p{2.8cm} >{\raggedright\arraybackslash}X >{\raggedright\arraybackslash}X@{}}
\toprule
\textbf{Item} & \textbf{Qwen2.5-VL-32B}\footnote{\url{https://huggingface.co/Qwen/Qwen2.5-VL-32B-Instruct}} & \textbf{InternVL3-14B}\footnote{\url{https://huggingface.co/OpenGVLab/InternVL3-14B-Instruct}} \\
\midrule
Decoder layers & 64 & 48 \\
Precision, attention & bfloat16, eager attention & bfloat16, eager attention \\
Decoding & greedy, repetition penalty 1.05 & greedy, no added repetition penalty \\
Output caps & CoT 512 tokens, Direct 24 tokens & as left \\
Image processing & resized to the visual-token budget & 448-pixel tiles, 256 visual tokens per tile \\
Visual-token budget & single-turn 256; follow-up, GQA, TallyQA 1280 & tiling under the same budgets \\
\bottomrule
\end{tabularx}
\end{minipage}
\end{table}

\clearpage
\begin{table}[h]
\centering
\footnotesize
\setlength{\tabcolsep}{3pt}
\renewcommand{\arraystretch}{1.12}
\caption{\textbf{Information conditions supplied as text.}
``The current answer'' is the answer to the Turn~2 question in follow-up counting.
Conditions that supply the required attributes are distinct from the label Sufficient given to generated prefixes (\appref{app:annotation}).}
\label{tab:app_a4}
\begin{tabularx}{\linewidth}{@{}>{\raggedright\arraybackslash}p{1.35cm} >{\raggedright\arraybackslash}p{2.0cm} >{\raggedright\arraybackslash}p{1.9cm} >{\raggedright\arraybackslash}X >{\raggedright\arraybackslash}p{2.2cm} >{\raggedright\arraybackslash}p{2.3cm}@{}}
\toprule
\textbf{Task} & \textbf{Condition} & \textbf{Position} & \textbf{Information supplied} & \textbf{States the current answer} & \textbf{Length control} \\
\midrule
Single-turn & None & -- & none & no & -- \\
\cmidrule(l{2pt}r{0pt}){2-6}
 & \hl{Complete} & user text before the question & color, shape, and location of every object & the count itself is not stated & -- \\
\cmidrule(l{2pt}r{0pt}){2-6}
 & Partial & as above & the same listing with the colors withheld & no & listing format shared except color \\
\cmidrule(l{2pt}r{0pt}){2-6}
 & Unrelated & as above & text unrelated to the image & no & token count matched to Complete \\
\midrule
Follow-up & \hl{Complete} & Turn~1 user text & color, shape, and quadrant of the objects & the count itself is not stated & -- \\
\cmidrule(l{2pt}r{0pt}){2-6}
 & Partial & as above & the attribute listing with the colors withheld & no & listing format shared except color \\
\cmidrule(l{2pt}r{0pt}){2-6}
 & Unrelated & as above & text unrelated to the image & no & token count matched to Complete line by line \\
\cmidrule(l{2pt}r{0pt}){2-6}
 & Label-only history & Turn~1 assistant reply & the correct quadrant name & only the Turn~1 answer & -- \\
\cmidrule(l{2pt}r{0pt}){2-6}
 & \hl{Attribute-stating history} & Turn~1 question and reply & the quadrant and the color and shape of its objects & the count itself is not stated & matched to Shape-only at 95 tokens \\
\cmidrule(l{2pt}r{0pt}){2-6}
 & Shape-only history & as above & the quadrant and the shapes of its objects, without colors & no & matched to Attribute-stating at 95 tokens \\
\midrule
GQA & None & -- & none & no & -- \\
\cmidrule(l{2pt}r{0pt}){2-6}
 & \hl{Complete} & Turn~1 user text & scene-graph listing, including the relation that identifies the target and the asked color & yes: contains the answer color (positive control) & -- \\
\cmidrule(l{2pt}r{0pt}){2-6}
 & Partial & as above & the scene information with the colors withheld & no & control without color information \\
\cmidrule(l{2pt}r{0pt}){2-6}
 & Unrelated & as above & text unrelated to the image & no & matched to Complete per tokenizer \\
\bottomrule
\end{tabularx}
\end{table}

\clearpage
\subsection{Prompts}
\label{app:prompts}

Boxes A.1--A.4 give the inputs used in the experiments; the English text of the prompts is reproduced verbatim, and square brackets mark task-dependent parts.

\begin{promptbox}[unbreakable,before skip=2pt,after skip=4pt,top=5pt,bottom=5pt]{Box A.1: Answer instructions (end of every evaluated question)}
\promptrole{CoT}
{\ttfamily\raggedright\frenchspacing\footnotesize
Inspect the image and reason step by step in at most 120 words using its visible evidence.
End with a plain line containing Answer: followed by exactly one \promptvar{option letter | nonnegative integer}.
Do not include \promptvar{option text | extra text} or Markdown on that line.\par}
\promptrole{Direct}
{\ttfamily\raggedright\frenchspacing\footnotesize
\promptvar{Choose one listed option.}
Give only the final answer without explanation.
End with a plain line containing Answer: followed by exactly one \promptvar{option letter | nonnegative integer}.
Do not include \promptvar{option text | extra text} or Markdown on that line.\par}
\promptrole{Immediate-answer control}
{\ttfamily\raggedright\frenchspacing\footnotesize
The CoT instruction verbatim; the assistant turn is teacher-forced to open with \texttt{Answer:} and decoding continues from there.\par}
\promptsep
\promptnote{Square brackets mark the task-dependent alternatives.
Each input is wrapped in the model's native chat template.}
\end{promptbox}

\begin{promptbox}{Box A.2: Single-turn input (image \texttt{multi\_val\_00000100}; answer C)}
\begin{minipage}[t]{0.22\linewidth}\vspace{0pt}
\includegraphics[width=\linewidth]{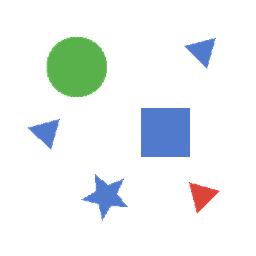}
\end{minipage}\hfill
\begin{minipage}[t]{0.75\linewidth}\vspace{0pt}
\promptrole{User}
\promptvar{image at left}
{\ttfamily\raggedright\frenchspacing\footnotesize
\promptvar{description; omitted under None}\\[3pt]
How many blue objects are there in the image?\\[3pt]
Options:\\
A. 3\\
B. 2\\
C. 4\\
D. 1\\
E. 5\\[3pt]
\promptvar{answer instruction from Box A.1}\par}
\end{minipage}
\promptsep
\promptnote{Complete lists \texttt{color}, \texttt{shape}, \texttt{size}, \texttt{location}, and \texttt{angle} for every object; Partial uses the same list with \texttt{color=not provided}; Unrelated is text unrelated to the image, matched to the token count of Complete.
The repeated attribute list is omitted here.}
\end{promptbox}

\begin{promptbox}{Box A.3: Follow-up input and histories (image arc-0012; the anchor is the red triangle in lower\_right; answer 1)}
\begin{minipage}[t]{0.22\linewidth}\vspace{0pt}
\includegraphics[width=\linewidth]{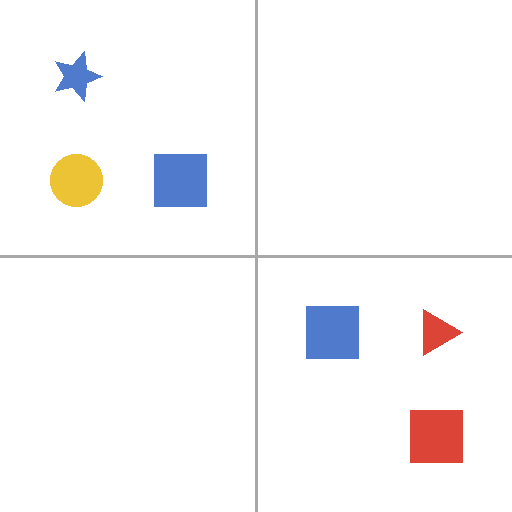}
\end{minipage}\hfill
\begin{minipage}[t]{0.75\linewidth}\vspace{0pt}
\promptrole{User, Turn~1}
\promptvar{image at left}
{\ttfamily\raggedright\frenchspacing\footnotesize
\promptvar{Additional information: description; omitted under None}\\[3pt]
The image is divided into four quadrants: upper\_left, upper\_right, lower\_left, lower\_right.\\
Which quadrant contains the only red triangle?\\
Reply with only the quadrant name -{}-{} one of upper\_left, upper\_right, lower\_left, lower\_right -{}-{} and nothing else.
Do not explain.\par}
\promptrole{Assistant, Turn~1 (teacher-forced)}
{\ttfamily\raggedright\frenchspacing\footnotesize
lower\_right\par}
\promptrole{User, Turn~2}
{\ttfamily\raggedright\frenchspacing\footnotesize
In that same quadrant, how many blue objects are there?\\[3pt]
\promptvar{answer instruction from Box A.1}\par}
\end{minipage}
\promptsep
\promptnote{Attribute-stating history: Turn~1 also asks for the objects of the quadrant by color and shape, and the fixed reply is}
{\ttfamily\raggedright\frenchspacing\footnotesize
lower\_right.
It contains a blue square, a red triangle, and a red square.\par}
\promptnote{Shape-only history: Turn~1 asks for the shapes in order, and the fixed reply is}
{\ttfamily\raggedright\frenchspacing\footnotesize
lower\_right.
It contains a visible square, a visible triangle, and a visible square.\par}
\end{promptbox}

\begin{promptbox}{Box A.4: Real-image input (GQA image 2379795, answer C; TallyQA image 2355775)}
\begin{minipage}[t]{0.22\linewidth}\vspace{0pt}
\includegraphics[width=\linewidth]{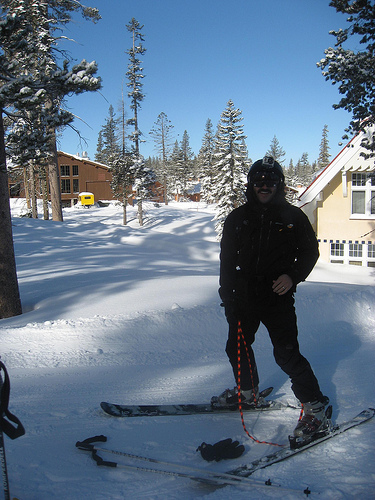}\\[4pt]
\includegraphics[width=\linewidth]{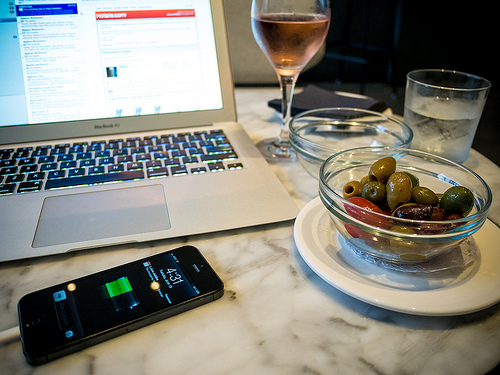}
\end{minipage}\hfill
\begin{minipage}[t]{0.75\linewidth}\vspace{0pt}
\promptrole{User, Turn~1}
\promptvar{image (top left)}
{\ttfamily\raggedright\frenchspacing\footnotesize
Keep this image available for my next question.
Reply only Ready.\\[3pt]
\promptvar{Additional information: description; omitted under None}\par}
\promptrole{Assistant, Turn~1 (fixed)}
{\ttfamily\raggedright\frenchspacing\footnotesize
Ready.\par}
\promptrole{User, Turn~2 (GQA example)}
{\ttfamily\raggedright\frenchspacing\footnotesize
What color is the rope that the skier is holding?\\[3pt]
Options:\\
A. red\\
B. brown\\
C. orange\\
D. white\\
E. purple\\[3pt]
\promptvar{answer instruction from Box A.1}\par}
\end{minipage}
\promptsep
\promptnote{The GQA Complete description contains the relation ``the skier is holding the rope'' and the attribute ``rope: orange''.
For TallyQA no description is added, and Turn~2 asks, for image 2355775, \texttt{How many laptops in the photo?} with a nonnegative-integer answer.}
\end{promptbox}

\clearpage
\subsection{Attribute-Expression Annotation and Intervention Stages}
\label{app:annotation}

The stages of \autoref{sec:expression} are defined by a fixed rule that judges how far the required attributes have been expressed in a generated prefix.
The rule consults the ground-truth attributes of the scene, so it is an oracle annotation for analysis, not a signal for access control at deployment.
It classifies visible text, and does not show that the model causally uses that text \citep{jacovi2020faithful}.

\begin{table}[h]
\centering
\footnotesize
\setlength{\tabcolsep}{3.5pt}
\renewcommand{\arraystretch}{1.15}
\caption{\textbf{Prefix categories and stage definitions.}
Categories are applied from top to bottom.
The follow-up answer is determined by the colors of the two non-anchor objects, and the anchor itself never has the asked color.
Objects are matched by color, and wrong shape names are tolerated.}
\label{tab:app_a5}
\textbf{Panel A: prefix categories}\\[2pt]
\begin{tabularx}{\linewidth}{@{}>{\raggedright\arraybackslash}p{2.3cm} >{\raggedright\arraybackslash}X >{\raggedright\arraybackslash}p{4.3cm}@{}}
\toprule
\textbf{Category} & \textbf{Criterion} & \textbf{Note} \\
\midrule
Incorrect & mentions a nonexistent object or a wrong color or count, or a closed enumeration implies a wrong count & judged before the other categories \\
\hl{Sufficient} & states the colors of both non-anchor objects, or completes a closed enumeration that implies the correct count & a closed enumeration is judged at the step that follows the last object mention \\
One color & states the color of exactly one non-anchor object & the other required color is not yet expressed \\
Uninformative & none of the above & includes mentions of the anchor alone \\
\bottomrule
\end{tabularx}

\medskip
\textbf{Panel B: intervention positions}\\[2pt]
\begin{tabularx}{\linewidth}{@{}>{\raggedright\arraybackslash}p{2.8cm} >{\raggedright\arraybackslash}X >{\raggedright\arraybackslash}X@{}}
\toprule
\textbf{Stage} & \textbf{Intervention position} & \textbf{Responses included} \\
\midrule
Before expression & the prediction of the token at which the first non-anchor color is written & responses in which this position is defined \\
After sufficient expression & the token after the prefix is first labeled Sufficient & responses that state at least one non-anchor color and reach a Sufficient prefix \\
\bottomrule
\end{tabularx}
\end{table}

The rule was developed on the 64-token prefixes of the independent set under both question-access conditions and fixed before any stage intervention ran.
The prefixes were inspected with answer correctness hidden, although correctness was visible for six images while the first version of the rule was written.
The judgment applied at each position has not been validated by an independent annotator, and the uncertainty of the cutoff positions is not included in the CIs of the results.

The before and after positions are not the moments just before and after one event, and several descriptive steps may separate them.
The comparison measures sensitivity at two content stages; it does not identify a switch at the moment of verbalization or a causal effect of verbalization.

\clearpage
\subsection{Fixed-Prefix Replay}
\label{app:replay}

The replay of Table~\ref{tab:prefix} uses the same initial text for each prefix source and changes only question-time access, and compares the additional knockout loss.

\begin{figure}[h]
\centering
\includegraphics[width=\linewidth]{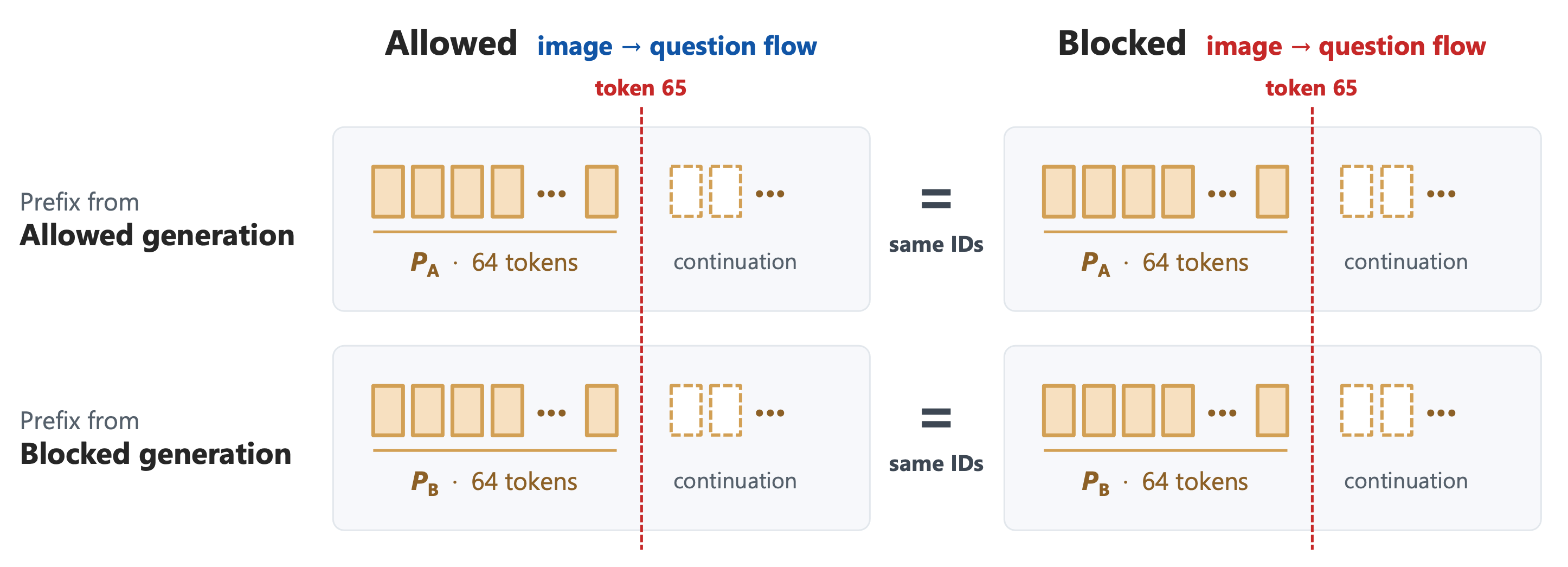}
\caption{\textbf{Fixed-prefix replay design.}
Rows indicate the source of the initial 64-token prefix; columns indicate question-time image access during replay.
Within each row, the prefix token IDs are identical across Allowed and Blocked conditions; prefixes in different rows need not match.
The image, Turn~1 cache, and current question are held fixed, while question-side and prefix states are recomputed under the replay condition.
Continuations may differ.
The dashed lines mark the onset of answer-side knockout at the prediction of token 65, with no layer restriction.
Each cell is compared against its own baseline without answer-side knockout under the same question-time access condition.}
\label{fig:app_a1}
\end{figure}

Replaying a prefix under the condition that generated it reproduced the original generation in all 400 checked continuations.
The comparison controls the initial visible text, but it fixes neither the internal states nor the text after token 64.

\begin{figure}[h]
\centering
\includegraphics[width=0.55\linewidth]{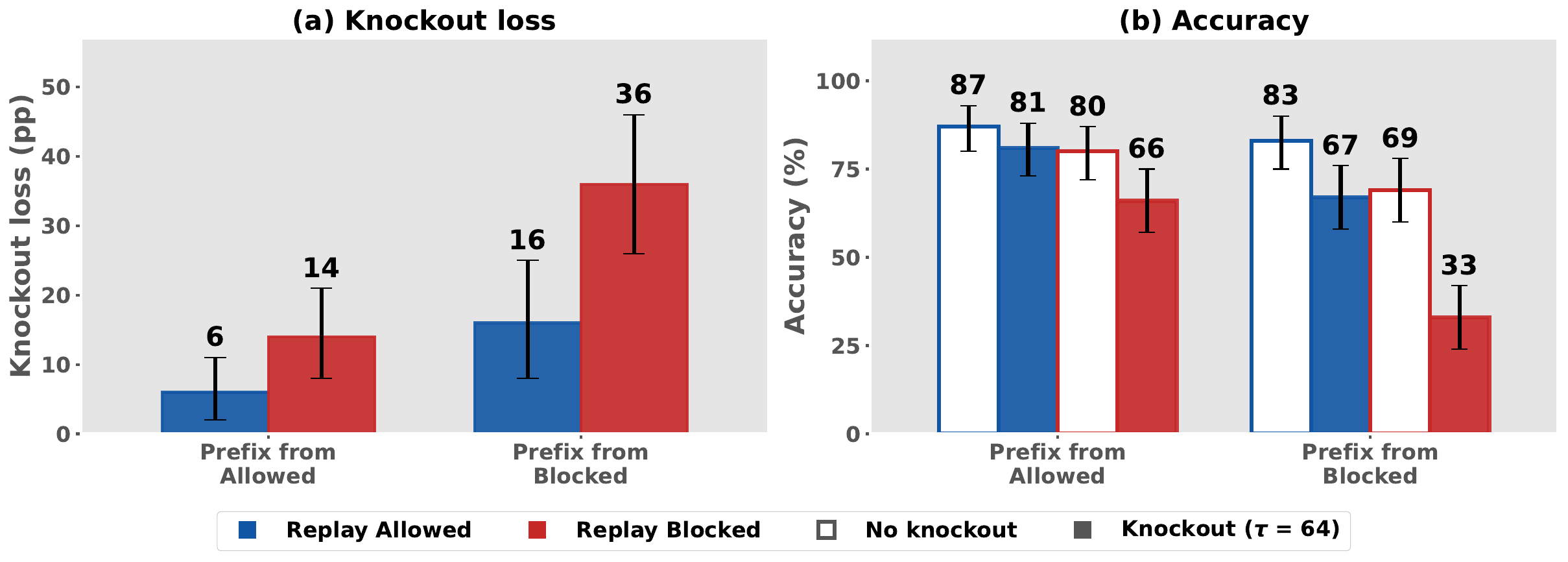}
\caption{\textbf{Knockout loss and accuracy in the four fixed-prefix replay conditions.} For both prefix sources, replay with question-time access Blocked loses more, but the baselines also differ. \textbf{(a)} Knockout loss and \textbf{(b)} baseline and knockout accuracy, grouped by the source of the 64-token prefix and colored by replay access. Independent set, Label-only history, CoT; within a source the same 64 token IDs are replayed with states recomputed, and knockout starts at the prediction of token 65 with no layer restriction.}
\label{fig:app_c3}
\end{figure}

\begin{promptbox}[unbreakable]{Box A.5: The same prefix, two continuations (image arc-0012; follow-up question with answer 1)}
\begin{minipage}[t]{0.2\linewidth}\vspace{0pt}
\includegraphics[width=\linewidth]{figures/appendix/example_arc0012.png}
\end{minipage}\hfill
\begin{minipage}[t]{0.77\linewidth}\vspace{0pt}
\promptrole{End of the shared prefix, generated with Allowed}
{\ttfamily\raggedright\frenchspacing\footnotesize
3. Examine the contents of the lower\_right quadrant:\par}
\promptrole{Continuation replayed with Allowed (excerpt): final answer 1}
{\ttfamily\raggedright\frenchspacing\footnotesize
\ \ - There is one red triangle.\\
\ \ - There is also one blue object, which appears to be a square.\par}
\promptrole{Continuation replayed with Blocked (excerpt): final answer 0}
{\ttfamily\raggedright\frenchspacing\footnotesize
\ \ - There is a red triangle.\\
\ \ - No other objects are present in this quadrant.\par}
\end{minipage}
\promptsep
\promptnote{Excerpts for readability; the replay uses the original 64 token IDs.
The example illustrates the operation; the results are in Table~\ref{tab:prefix} and \autoref{fig:app_c3}.}
\end{promptbox}

\clearpage
\subsection{Counterfactual Image-KV Swap and Implementation Checks}
\label{app:swap}

In \autoref{fig:expression} (right), the prefix computed from the recipient image is kept, and the KV at the image positions is replaced by KV computed from another image.
The donor KV comes from a separate full-access prefill of the same Turn~1, and the keys and values at the image positions of every layer are replaced.
The KV at the text positions of Turn~1, the current question, and the generated prefix keeps the values computed from the original image.
The operation therefore does not re-read an image and update every state; it combines states derived from different images.

\begin{table}[h]
\centering
\footnotesize
\setlength{\tabcolsep}{3pt}
\renewcommand{\arraystretch}{1.15}
\caption{\textbf{Swap conditions and implementation checks.}
In Panel~A, $c$ is the original correct count; donors A and B have the smaller and the larger value of $\{0,1,2\}\setminus\{c\}$ as their correct count, which need not be smaller or larger than $c$.}
\label{tab:app_a6}
\textbf{Panel A: interventions and controls}\\[2pt]
\begin{tabularx}{\linewidth}{@{}>{\raggedright\arraybackslash}p{1.9cm} >{\raggedright\arraybackslash}p{2.9cm} >{\raggedright\arraybackslash}p{2.6cm} >{\raggedright\arraybackslash}p{1.7cm} >{\raggedright\arraybackslash}X@{}}
\toprule
\textbf{Condition} & \textbf{Image change} & \textbf{Image KV} & \textbf{Text KV} & \textbf{Purpose} \\
\midrule
No swap & none & original KV kept & kept & baseline of the forced-prefix continuation \\
Self swap & the same image & replaced by KV computed from the same image & kept & consistency of the replacement itself \\
\hl{Target A} & objects in the anchor quadrant recolored & replaced by donor A's KV & from the original image & directional response to another correct count \\
\hl{Target B} & objects in the anchor quadrant recolored & replaced by donor B's KV & from the original image & response to the other correct count \\
Unrelated swap & one object in the decoy quadrant recolored & replaced by the changed image's KV & from the original image & control for an image change that keeps the count \\
\bottomrule
\end{tabularx}

\medskip
\textbf{Panel B: implementation and execution-path checks}\\[2pt]
\begin{tabularx}{\linewidth}{@{}>{\raggedright\arraybackslash}p{2.7cm} >{\raggedright\arraybackslash}p{3.3cm} >{\raggedright\arraybackslash}p{3.2cm} >{\raggedright\arraybackslash}X@{}}
\toprule
\textbf{Check} & \textbf{Runs compared} & \textbf{Items} & \textbf{Result} \\
\midrule
Cache parity & cached vs uncached full-access generation & 211 & token sequences identical in 211/211 \\
Fixed-prefix parity & fixed-prefix vs native generation & 6 & identical in 6/6 \\
Own-condition replay & own-condition replay vs free generation & 400 continuations & identical in 400/400 \\
Stage-prefix match & prefix up to the stage intervention vs free prefix & before: Allowed 96, Blocked 93; after: Allowed 85, Blocked 72 & the pre-intervention prefix matches in all \\
Self-swap parity & self swap vs no swap & Allowed 30, Blocked 29 & token sequences identical in all, both conditions \\
No-swap vs free path & no swap vs free generation of the same run & Allowed 96, Blocked 93 & token sequences identical 68/96 and 74/93; final answers identical 93/96 and 92/93 \\
\bottomrule
\end{tabularx}
\end{table}

Swap comparisons use the no-swap continuation of the same arm and run as the reference, and do not assume exact identity with free generation.
Each run checks the application position, the prefix reproduction, the replaced KV, the planned item set, format violations, and output caps.
Target swaps are evaluated before expression in both question-access conditions and after sufficient expression in the Allowed condition.

\clearpage
\section{Model Family Ablations (InternVL3-14B)}
\label{app:models}
\label{app:internvl}

To check the correspondence with Qwen2.5-VL-32B in the main text, we compare response formats, description content, and histories with question-time access on the synthetic tasks in InternVL3-14B \citep{zhu2025internvl3}.
All results in this appendix use InternVL3-14B on the synthetic counting tasks, with the same intervention definitions and outcome measures as in the main text.
A change of model family also changes size, training, and implementation, so these comparisons do not isolate a single factor.

\subsection{Response Format}
\label{app:internvl_format}

\begin{figure}[h]
\centering
\includegraphics[width=0.85\linewidth]{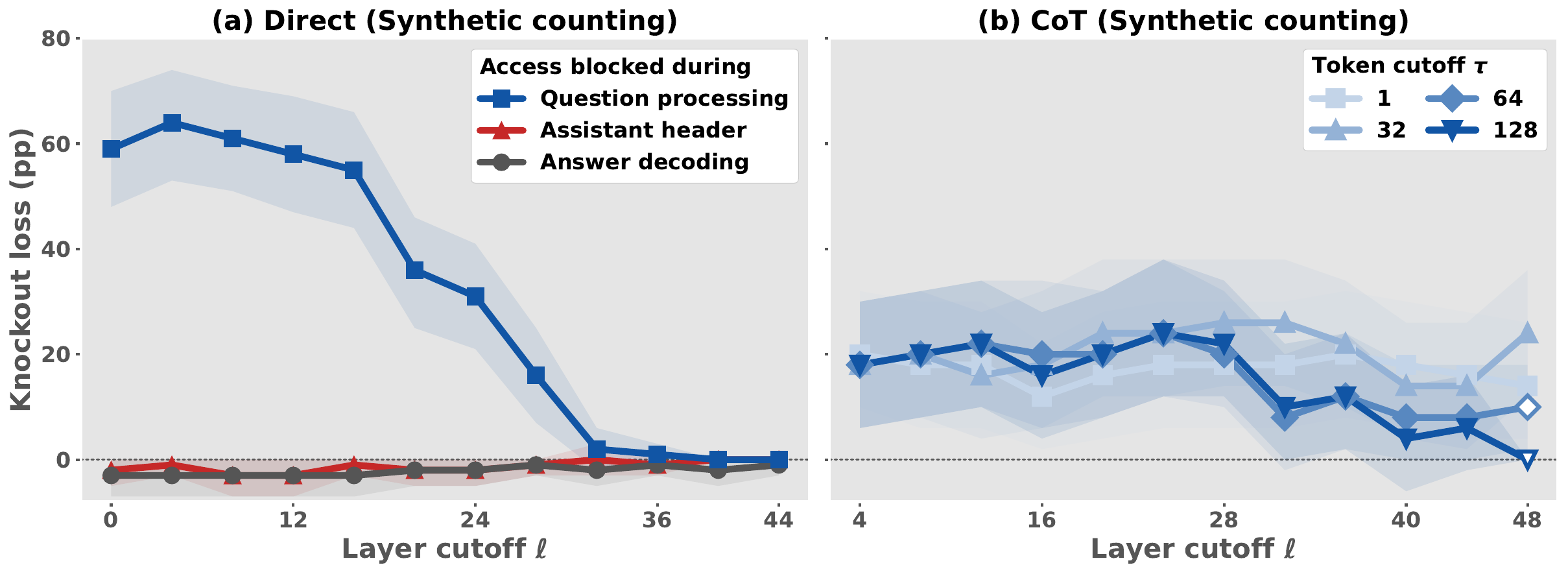}
\caption{\textbf{Comparison of knockout loss across response formats.} Direct answers lose accuracy when the question rows are severed but little otherwise, whereas CoT answers also lose accuracy under answer-side knockout. \textbf{(a)} Direct: question, assistant-header, or decode rows severed separately at layers $k\ge\ell$. \textbf{(b)} CoT: answer-side layer and token cutoffs combined, with question-time access Allowed.}
\label{fig:app_b1}
\end{figure}

For Direct answers, severing image access of the question rows from shallow layers causes a loss, whereas severing the header or decode rows produces little observed loss.
For CoT, answer-side knockout causes a loss.
This contrast supports the main-text result for Qwen2.5-VL-32B that Direct and CoT answers are sensitive through different access routes.

\subsection{Description Type}
\label{app:internvl_description}

\begin{figure}[h]
\centering
\includegraphics[width=0.85\linewidth]{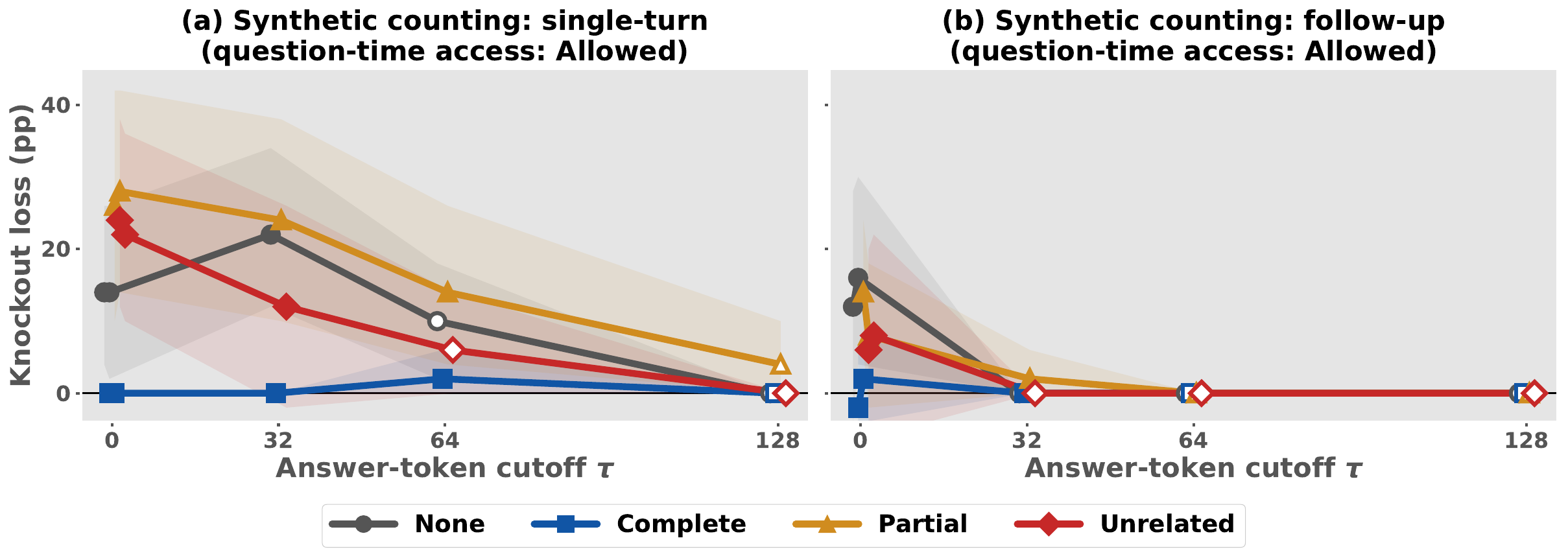}
\caption{\textbf{Comparison of knockout loss across description conditions.} Complete descriptions keep loss small in single-turn counting, where Partial and Unrelated do not; in follow-up counting the conditions separate less. \textbf{(a)} Single-turn and \textbf{(b)} follow-up counting with the Label-only history; CoT, question-time access Allowed, no layer restriction.}
\label{fig:app_b2}
\end{figure}

In single-turn counting, Complete, which supplies the required colors, shows a small loss, whereas Partial, which withholds the colors, and unrelated text do not remove the loss at answer onset.
This supports the main-text result that containing the attributes the answer needs matters, not merely the presence of added text.
In follow-up counting (\autoref{fig:app_b2}(b)), the differences between conditions are small, and the separation seen in single-turn counting does not appear.

\clearpage
\subsection{History Content and Question-Time Access}
\label{app:internvl_history}

\begin{figure}[h]
\centering
\includegraphics[width=0.5\linewidth]{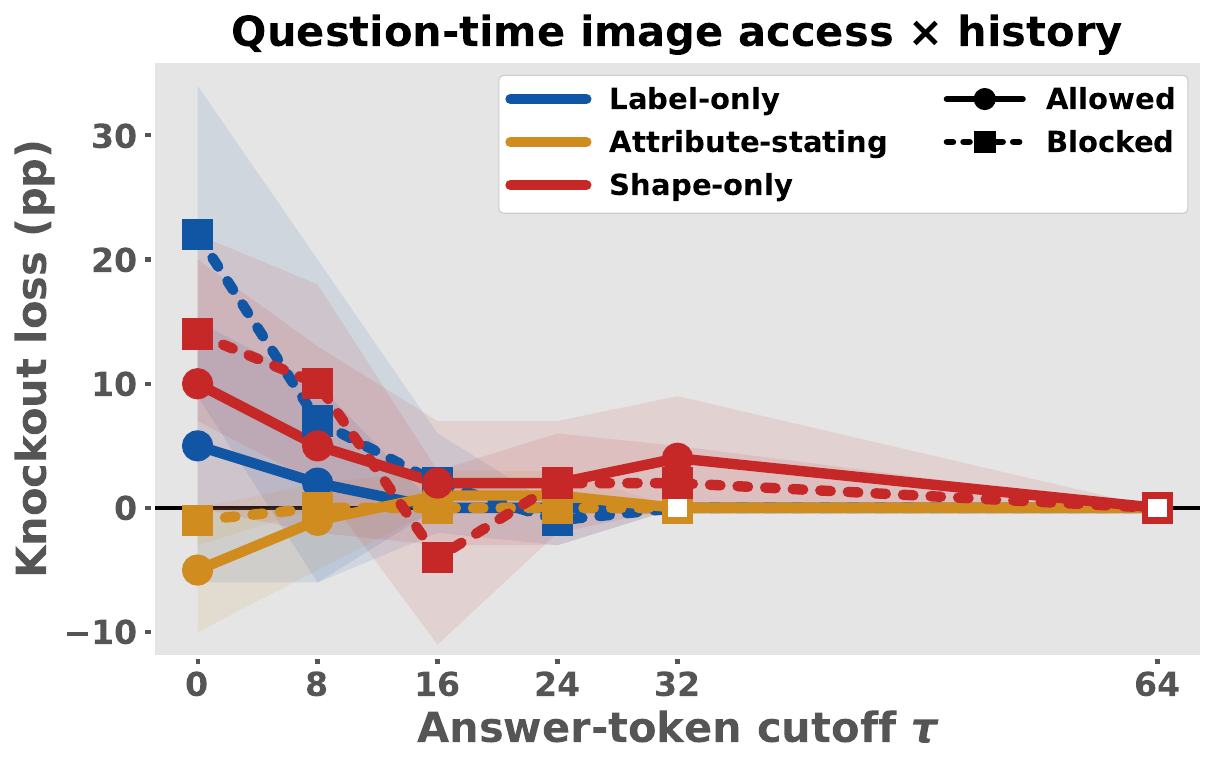}
\caption{\textbf{Comparison of knockout loss across histories and question-time access.} Attribute-stating histories keep loss near zero; the Label-only access difference appears at answer onset but not at later cutoffs. Follow-up counting, CoT, no layer restriction; colors give histories and line styles access (solid: Allowed; dashed: Blocked), each against its own baseline.}
\label{fig:app_b3}
\end{figure}

With the Attribute-stating history, which contains the required attributes, the loss is small in both access conditions, and the answer-onset loss seen with the Label-only and Shape-only histories is suppressed.
This supports the main-text comparison that separates the correctness of an earlier answer from the sufficiency of the information for the current question.

With the Label-only history, the answer-onset loss is larger under Blocked than under Allowed, so the question-access difference has the same direction as in Qwen2.5-VL-32B.
The difference becomes small at later cutoffs, however, and the persistent difference seen in Qwen2.5-VL-32B does not reproduce.
The effect of question-time access on the sensitivity of later generation therefore depends on the model and on the generation position.

\clearpage
\section{Supplementary Controls}
\label{app:controls}

\subsection{Immediate-Answer Control and Independent Confirmation}
\label{app:format_controls}

The immediate-answer control keeps the CoT instruction and tests whether the format difference of \autoref{sec:format} arises from the instruction text alone.

\begin{figure}[h]
\centering
\includegraphics[width=0.8\linewidth]{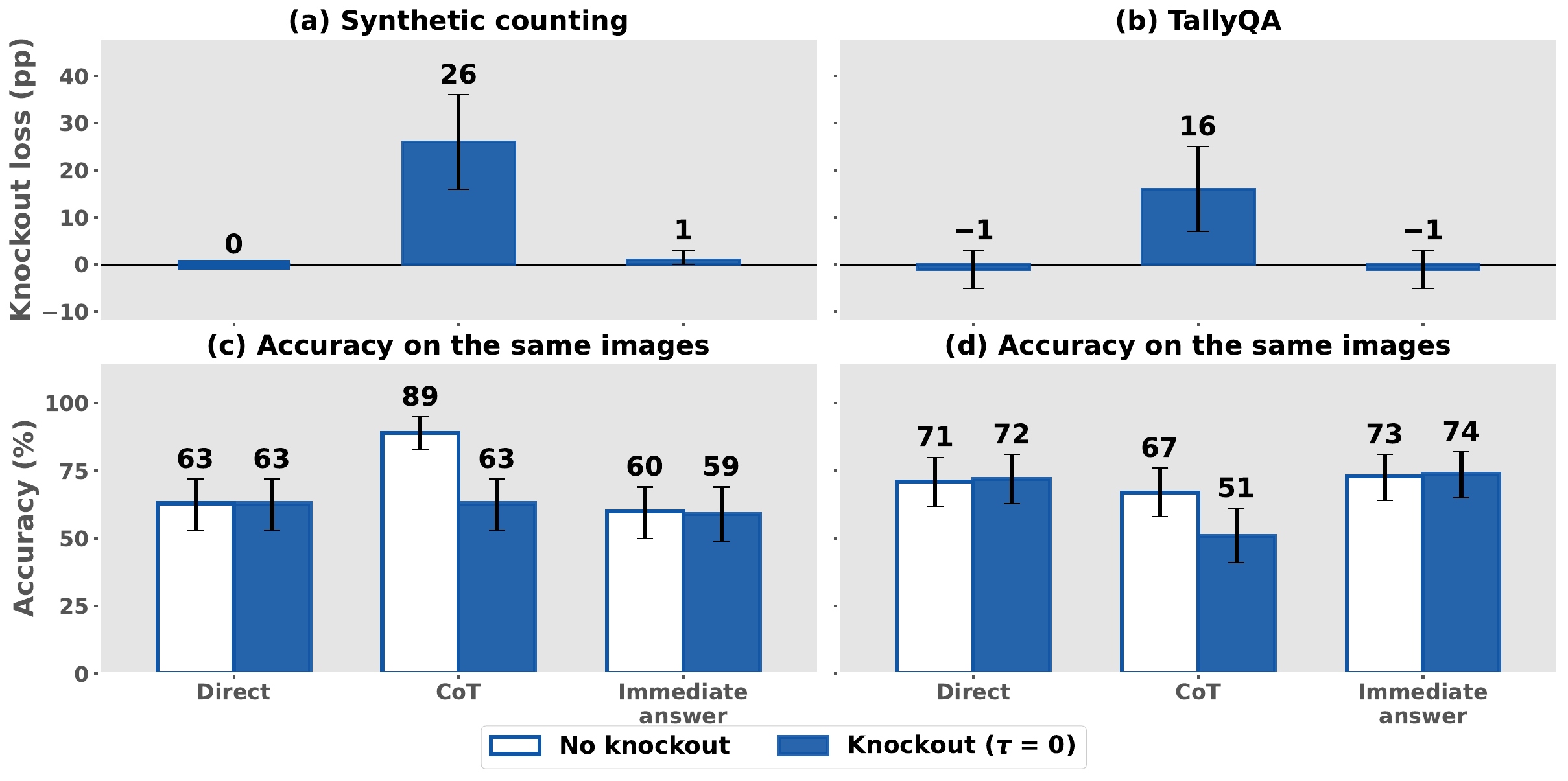}
\caption{\textbf{Comparison of knockout loss and accuracy across response formats, including the immediate-answer control.} The immediate-answer control, which keeps the CoT instruction, shows no CoT-like loss. \textbf{(a, c)} Count-balanced single-turn fresh set and \textbf{(b, d)} TallyQA; \textbf{(a, b)} knockout loss and \textbf{(c, d)} baseline and knockout accuracy. Question-time access Allowed, $\tau=0$, no layer restriction.}
\label{fig:app_c1}
\end{figure}

The immediate-answer control does not show the CoT loss, so the instruction text alone does not explain the format difference.
The three formats also differ in response length, so the comparison does not separate the content of reasoning from the effect of intervening on a long generation.

\subsection{Additional Information Conditions}
\label{app:desc_history}

Question-access conditions not shown in \autoref{fig:desc_synthetic} complement the relation between description content and loss.

\begin{figure}[h]
\centering
\includegraphics[width=\linewidth]{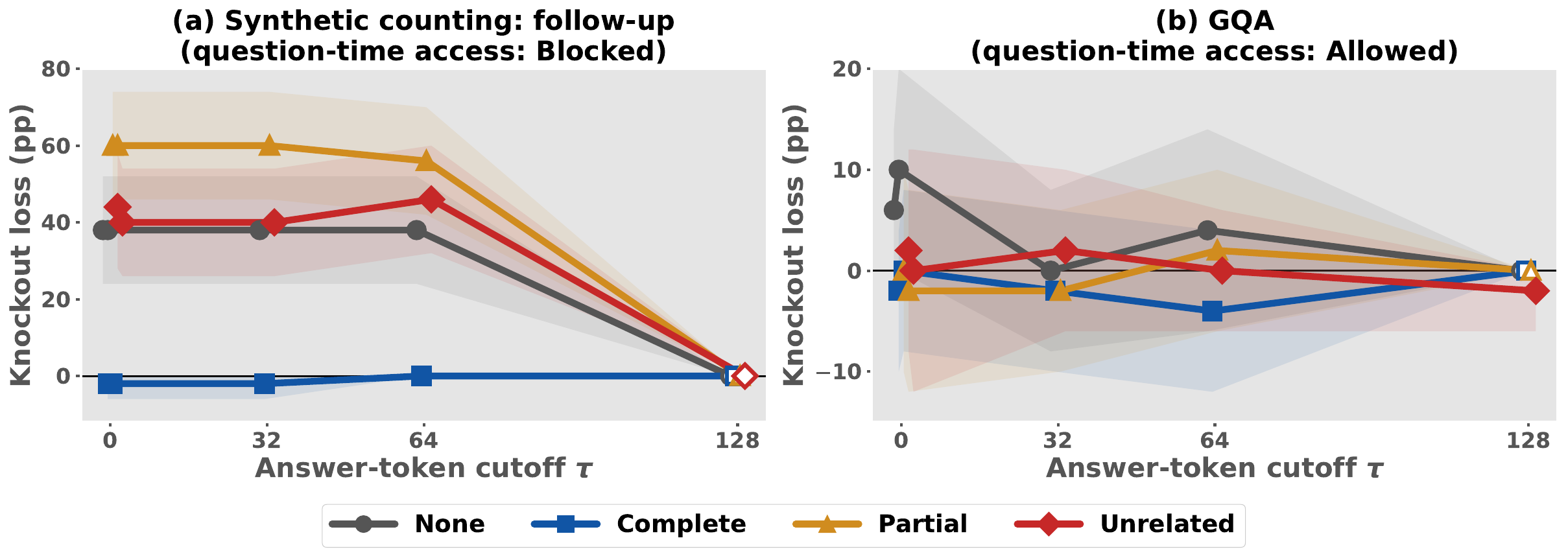}
\caption{\textbf{Comparison of knockout loss across description conditions in the complementary access conditions.} Complete keeps loss small in follow-up counting under Blocked; on GQA under Allowed the loss is small even without a description. \textbf{(a)} Follow-up counting with the Label-only history, question-time access Blocked; \textbf{(b)} GQA, question-time access Allowed; CoT, no layer restriction.}
\label{fig:app_c2}
\end{figure}

In follow-up counting under Blocked, Complete also keeps the loss small, whereas descriptions that withhold the required colors or are unrelated do not.
On GQA under Allowed, the loss is small even without a description, unlike the Blocked condition of the main text.

\clearpage
\subsection{The Conditions behind the Fixed-Prefix Replay}
\label{app:qaccess_prefix}

The differences in Table~\ref{tab:prefix} are split here into the loss and accuracy of the four underlying replay conditions.

For both prefix sources the replay with question-time access Blocked loses more.
The baseline accuracy also differs between conditions, so the difference in final knocked-out accuracy is not read as the question-access difference in loss.
The two source-specific differences are separate comparisons, not an additive decomposition of the free-generation difference.
The knocked-out Blocked continuations show a recurring failure pattern: on the profile set every knocked-out Blocked response at $\tau\le32$ answers ``0'', and on the independent set 43 of the 67 errors at $\tau=64$ state that only the Turn~1 object is present.

\clearpage
\subsection{Attribute Stages and Common Sets}
\label{app:stage_more}

\autoref{tab:app_c1} gives the responses that enter the before and after comparisons of \autoref{fig:expression}.
Because the stages are selected from the generated content, conditioning on reaching them can change the evaluated population \citep{hernan2020causal}; a common set aligns the images but not the elapsed generation or the difficulty.

\begin{table}[h]
\centering
\footnotesize
\setlength{\tabcolsep}{4pt}
\renewcommand{\arraystretch}{1.15}
\caption{\textbf{Populations of the stage analyses.}
Paired comparisons within a row use the same images; different rows need not share a population.}
\label{tab:app_c1}
\begin{tabularx}{\linewidth}{@{}>{\raggedright\arraybackslash}p{3.6cm} >{\raggedright\arraybackslash}p{2.1cm} >{\raggedright\arraybackslash}X r@{}}
\toprule
\textbf{Analysis} & \textbf{Question access} & \textbf{Selection} & \textbf{$N$} \\
\midrule
\hl{Stage knockout (main text)} & Allowed & both before and after positions are defined & 85 \\
\hl{Stage swap (main text)} & Allowed & Uninformative prefix at the before position, and the after position is evaluable & 79 \\
Stage knockout, both arms & Allowed / Blocked & before and after are defined in both arms & 66 \\
Fixed-prefix replay & both & 64-token prefixes comparable under both replay conditions & 100 \\
Token-64 content categories & both & 64-token prefixes of generation without knockout & 100 each \\
\bottomrule
\end{tabularx}
\end{table}

\begin{figure}[h]
\centering
\includegraphics[width=0.5\linewidth]{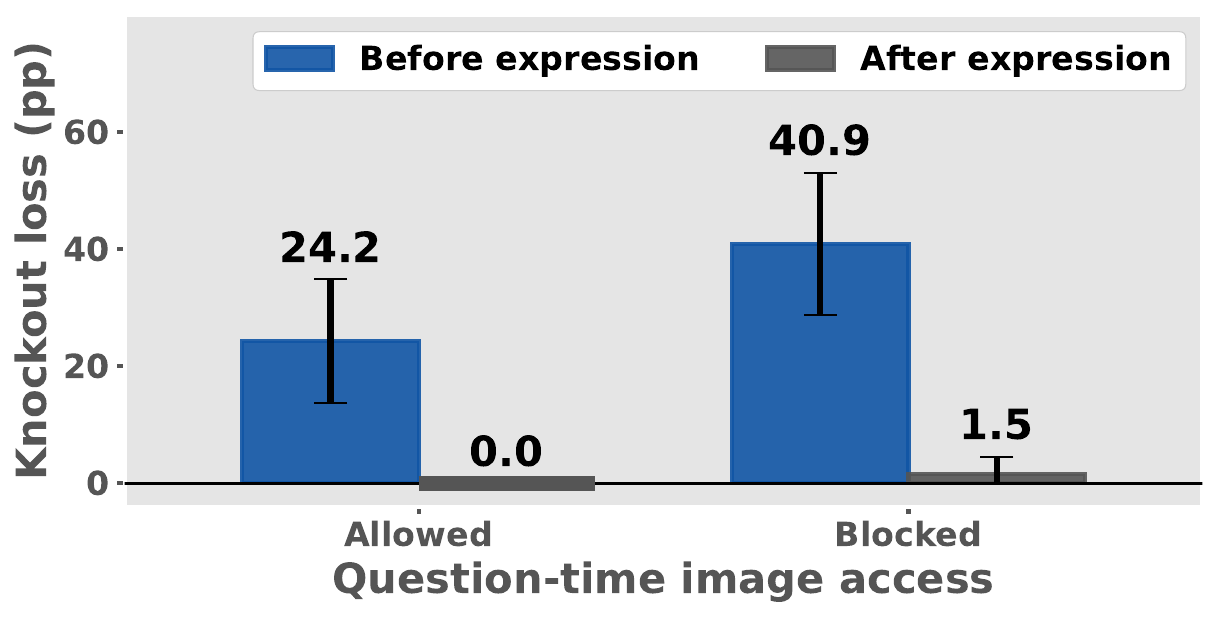}
\caption{\textbf{Comparison of knockout loss before and after sufficient expression on the common set.} The before--after difference holds on the images that reach both stages in both access conditions. Independent set, Label-only history, CoT; each response's intervention positions come from its own free generation.}
\label{fig:app_c4}
\end{figure}

The before--after difference survives the restriction to the same images in both access conditions, and all eligible responses give the same ordering (Allowed 16.7 vs 0.0~pp; Blocked 29.0 vs 1.4~pp).
What the restriction cannot remove is that the two stages also differ in elapsed generation and answer commitment, which is why the comparison does not establish a causal effect of writing the attributes.

\begin{figure}[h]
\centering
\includegraphics[width=0.7\linewidth]{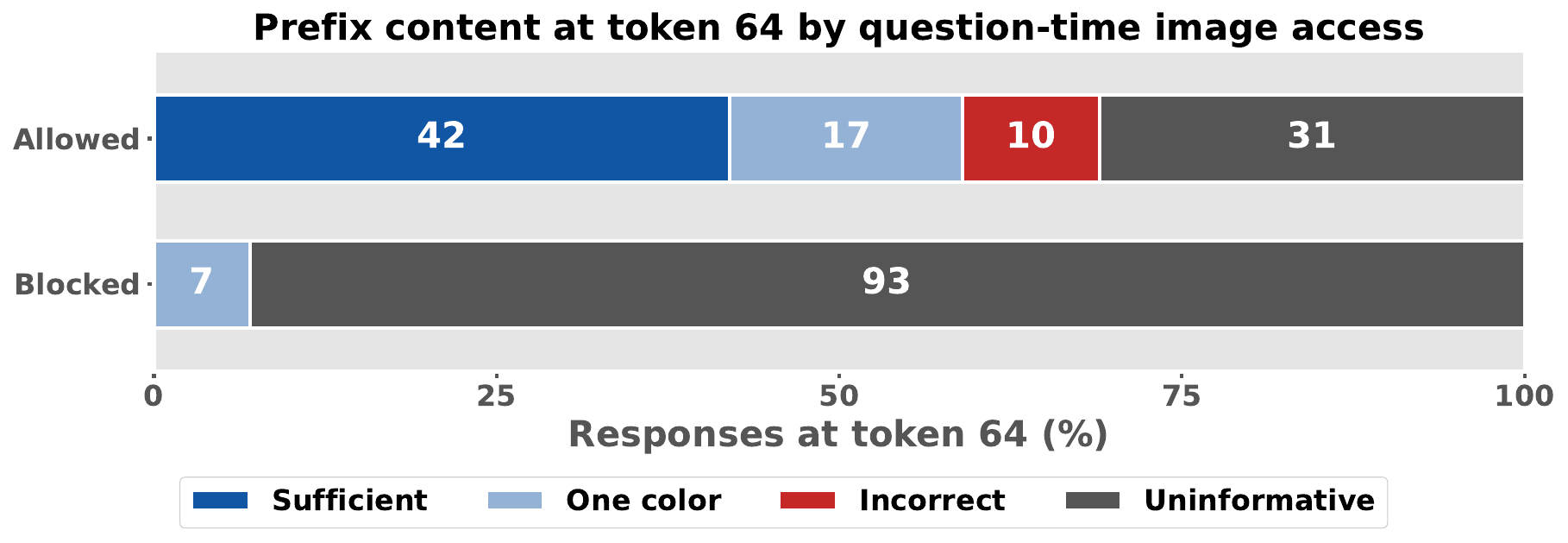}
\caption{\textbf{Share of responses by prefix category at token 64.} At the same token position, Allowed and Blocked responses have expressed different content. Free generations on the independent set, classified by \autoref{tab:app_a5}; a descriptive comparison, not a manipulation of attribute expression.}
\label{fig:app_c5}
\end{figure}

The same token position thus falls at different content stages under the two access conditions, and the knockout loss at that position follows the content already expressed: Sufficient prefixes keep their accuracy (Allowed 42 $\to$ 42 correct), whereas Uninformative prefixes lose it (Allowed 27 $\to$ 23; Blocked 62 $\to$ 29).
This is why the stage analyses align the cutoff to content rather than to token position.

\clearpage
\subsection{Counterfactual Swap Controls}
\label{app:swap_controls}

The change in donor-answer rate of \autoref{fig:expression} (right) is split into the raw rates without a swap and the response to an image change unrelated to the task.

\begin{figure}[h]
\centering
\includegraphics[width=\linewidth]{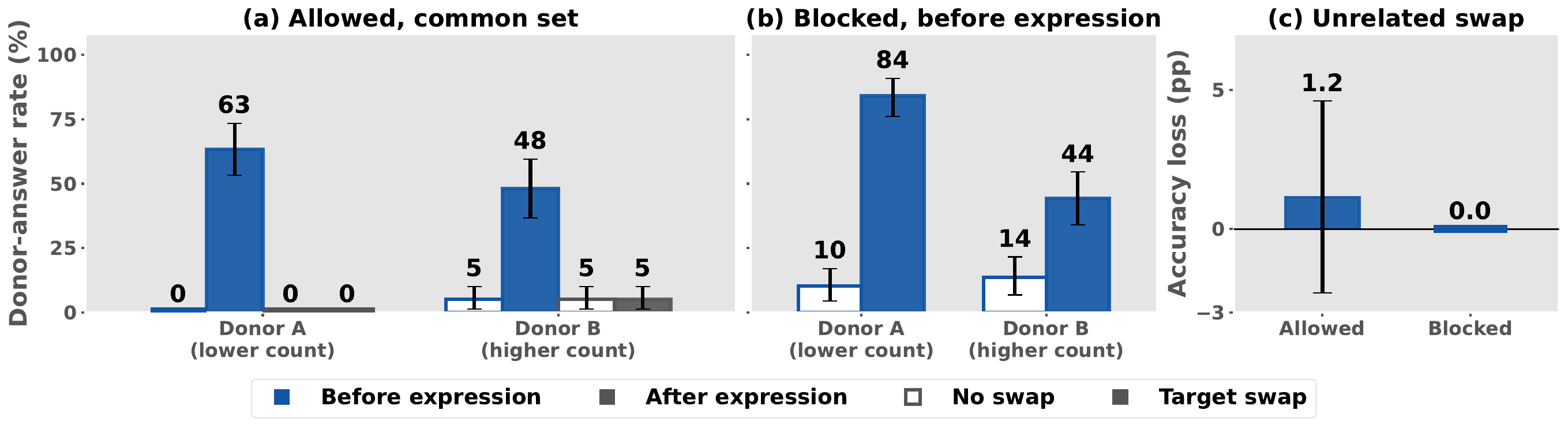}
\caption{\textbf{Donor-answer rates and unrelated-swap accuracy loss under counterfactual image-KV swap.} Target swaps raise the donor-answer rate before expression but not after, and the unrelated swap costs little accuracy. \textbf{(a)} Donor-answer rate under no swap and target swap on the common set, Allowed, before and after expression; \textbf{(b)} the same before expression under Blocked, on images with an Uninformative prefix; \textbf{(c)} accuracy loss under the unrelated swap. Rates are in \%, accuracy differences in pp; the self-swap checks are in \autoref{tab:app_a6}.}
\label{fig:app_c6}
\end{figure}

After expression, donor B's answer also occurs without a swap, so the rate after a target swap alone does not measure the effect of the replacement; the main text reports the difference from no swap on the same images.
The unrelated swap changes accuracy little (Allowed 1.2 [$-2.3$, 4.6]~pp; Blocked 0.0~pp), so the directional response to target swaps before expression is hard to capture with an account in which any image-KV replacement disrupts answers uniformly.

The self and unrelated swaps do not exclude every effect of a mixed KV state.
The results are read as sensitivity to the specified replacement, not as the share of a pathway used in ordinary generation or as evidence that image information is lost from the internal states after expression.

\end{document}